\documentclass[11pt]{article}
\usepackage[preprint]{acl}
\usepackage{times}
\usepackage{latexsym}
\usepackage[T1]{fontenc}
\usepackage[utf8]{inputenc}
\usepackage{microtype}
\usepackage{courier}
\usepackage{graphicx}
\usepackage{multirow}
\usepackage{algorithm}
\usepackage{algorithmic}
\usepackage{booktabs}
\usepackage{amsmath,amssymb}
\allowdisplaybreaks[1]
\newsavebox{\displaybox}
\newcommand{\fitdisplay}[1]{\sbox{\displaybox}{$\displaystyle #1$}\ifdim\wd\displaybox>\columnwidth\resizebox{\columnwidth}{!}{\usebox{\displaybox}}\else\usebox{\displaybox}\fi}
\title{GARRET: Generative Feedback for Region Retrieval\\in Visual Documents}
\author{
  Weijian Jian\textsuperscript{1}\thanks{Equal contribution.},
  Siyang Chen\textsuperscript{2,3,1*},
  Yuguang Yang\textsuperscript{4,1},
  Yajun Zhang\textsuperscript{1}\\
  \bfseries Chunyu Xie\textsuperscript{1},
  Dawei Leng\textsuperscript{1}\thanks{Corresponding Author, E-mail: \texttt{lengdawei@360.cn}},
  Yuhui Yin\textsuperscript{1}\\[5pt]
  {\normalfont\normalsize
    \textsuperscript{1}360 AI Research\quad
    \textsuperscript{2}SAIS, UCAS\quad
    \textsuperscript{3}CASIA\quad
    \textsuperscript{4}SEIE, BUAA}
}
\begin{document}
\maketitle

\begin{abstract}
Multimodal retrieval-augmented generation over visually rich documents requires evidence at a finer granularity than entire pages. Existing visual retrievers typically index or return full pages, whereas training region-level retrievers often requires costly query-to-region annotations or noisy model-generated labels. We introduce GARRET, a region-retrieval framework trained from (page, question, answer) triplets through differentiable generative feedback. A shared context-preserving retriever produces page and region multi-vector representations. During training, their retrieval scores define a soft routing distribution over candidate region crops. An auxiliary vision-language model attends to each crop separately, aggregates crop-specific contexts using the routing weights, and predicts the question followed by the answer under a next-token prediction objective. Gradients through the routing distribution assign query-conditioned credit to the indexed region representations, without requiring query-to-region annotations. At inference, the auxiliary model is removed. GARRET also reuses the contextualized region representations, together with page summary tokens, to construct a compact page representation. We introduce \textsc{ViDoRe-V3-Region}, a corpus-wide benchmark containing 200,682 fixed candidate regions across eight domains. GARRET achieves a macro nDCG@10 of 0.4126, outperforming the strongest visual baseline by 2.94 points. On page retrieval, it remains comparable to ColQwen2.5-v0.2 while indexing 47.8\% fewer vectors per page.

\end{abstract}

\begin{figure*}[t]
    \centering
    \includegraphics[width=\textwidth]{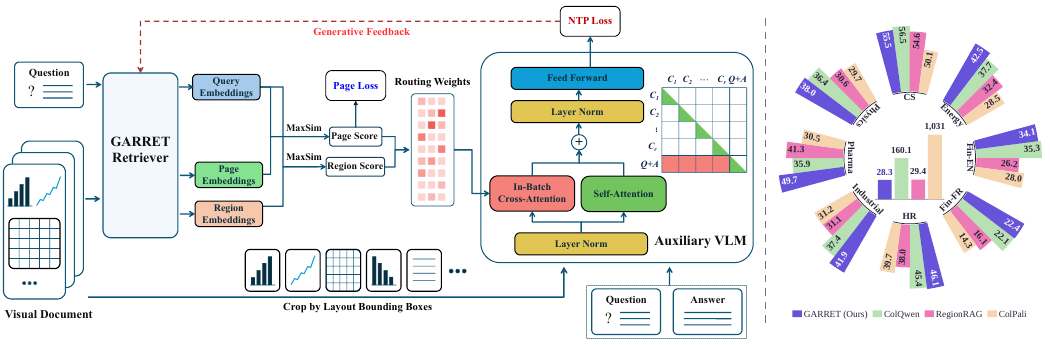}
    \caption{Overview of the GARRET framework and its performance. \textbf{Left:} During training, the retriever computes routing weights for candidate regions. An auxiliary VLM uses these weights via in-batch cross-attention to process isolated visual crops. By predicting the paired question and answer, the VLM backpropagates a next-token prediction (NTP) loss as generative feedback, teaching the retriever latent region relevance. \textbf{Right:} GARRET achieves state-of-the-art region retrieval across eight domains in ViDoRe V3 while requiring significantly fewer indexed vectors than existing baselines.}
    \label{fig:method}
\end{figure*}

\section{Introduction}
\label{sec:introduction}

Retrieval-augmented generation (RAG) has naturally expanded to visually rich documents, where critical information resides in layouts, tables, and figures \citep{yu2024visrag,tanaka2025vdocrag}. Recent visual retrievers, such as ColPali and ColQwen \citep{faysse2024colpali}, directly encode rendered pages to bypass error-prone OCR pipelines, achieving strong page-level performance. However, retrieving entire pages introduces severe redundancy. The actual evidence for a query is often confined to a short paragraph or a small figure. Feeding full pages into a generation model increases computational costs, introduces irrelevant context, and exacerbates hallucination risks. Thus, multimodal RAG demands a finer granularity: retrieving the exact supporting region.

Developing robust region-level retrievers, however, faces significant hurdles. Dynamic patch-grouping methods like RegionRAG \citep{li2026regionrag} construct regions on the fly, precluding efficient offline indexing. Conversely, block-based retrievers like LFRAG \citep{zhu2026lfraglayoutorientedfinegrainedretrievalaugmented} support offline indexing but require prohibitively expensive query-to-region annotations. Although large vision-language models (VLMs) can generate pseudo-labels, this process is computationally expensive and often yields noisy supervision. Thus, the lack of scalable, high-quality training data remains the primary bottleneck for fine-grained visual retrieval.

To overcome this bottleneck, we leverage the abundant $(\text{page}, \text{question}, \text{answer})$ triplets already available in existing datasets. While the exact bounding box is unannotated, the \textit{answer} provides an implicit supervisory signal: a region is relevant if its visual content enables a model to generate the correct answer. Based on this insight, we propose GARRET (\textbf{G}enerative \textbf{A}nswer feedback for \textbf{R}egion \textbf{RET}rieval), a framework that treats region relevance as a latent variable optimized directly via generative feedback. 

During training, GARRET extracts query-independent layout regions \citep{wang2024mineru} and computes their relevance scores to form a soft routing distribution. An auxiliary VLM then processes these isolated visual crops via retriever-weighted in-batch cross-attention. By training the VLM to jointly predict the paired question and answer using a next-token prediction (NTP) objective, the loss backpropagates directly through the routing scores. This mechanism naturally rewards regions that support generation and penalizes distractors, without requiring manual query-to-region training annotations or fixed VLM pseudo-labels. Post-training, the auxiliary VLM is discarded, yielding a standalone, highly effective retriever.

To ensure regions retain their semantic background, GARRET introduces a context-preserving multi-granularity embedding architecture, a visual analogue to late chunking \citep{gunther2024latechunking,yan2026visuallatechunking}. Region tokens are appended to the base page tokens, allowing each region to absorb global page context while blocking cross-region attention. Crucially, during offline indexing, the base page embeddings are discarded. Only the contextualized region embeddings and a small set of page-summary embeddings are retained. This design compresses dense page information into a highly compact, shared index that efficiently supports both independent region retrieval and full-page retrieval.

Finally, to rigorously evaluate fine-grained retrieval, we construct a novel benchmark spanning eight domains from ViDoRe V3 \citep{loison2026vidorev3}, comprising 200,682 fixed candidate regions with human-verified positive relevance judgments for corpus-level evaluation. On this benchmark, GARRET achieves an nDCG@10 of 0.4126, establishing a new state-of-the-art by a substantial margin. Furthermore, GARRET matches the page-level performance of the ColQwen family while requiring $47.8\%$ fewer indexed vectors, demonstrating exceptional storage efficiency.

Our main contributions are summarized as follows:
\begin{itemize}
    \item We propose a weakly supervised training framework that learns fine-grained visual retrieval directly from $(\text{page}, \text{question}, \text{answer})$ triplets. By utilizing an auxiliary VLM to provide differentiable generative feedback, GARRET eliminates the reliance on manual query-to-region annotations.
    \item We introduce a context-preserving multi-granularity architecture (visual late chunking). It enables regions to absorb global page context and discards dense page tokens during indexing, creating a highly compact representation for both region- and page-level retrieval.
    \item We construct a comprehensive region-level retrieval benchmark with over 200K fixed candidate regions. Extensive evaluations demonstrate that GARRET achieves state-of-the-art region retrieval and matches leading page-level retrievers while reducing index size by $47.8\%$.
\end{itemize}

\begin{figure}[t]
    \centering
    \includegraphics[width=\columnwidth]{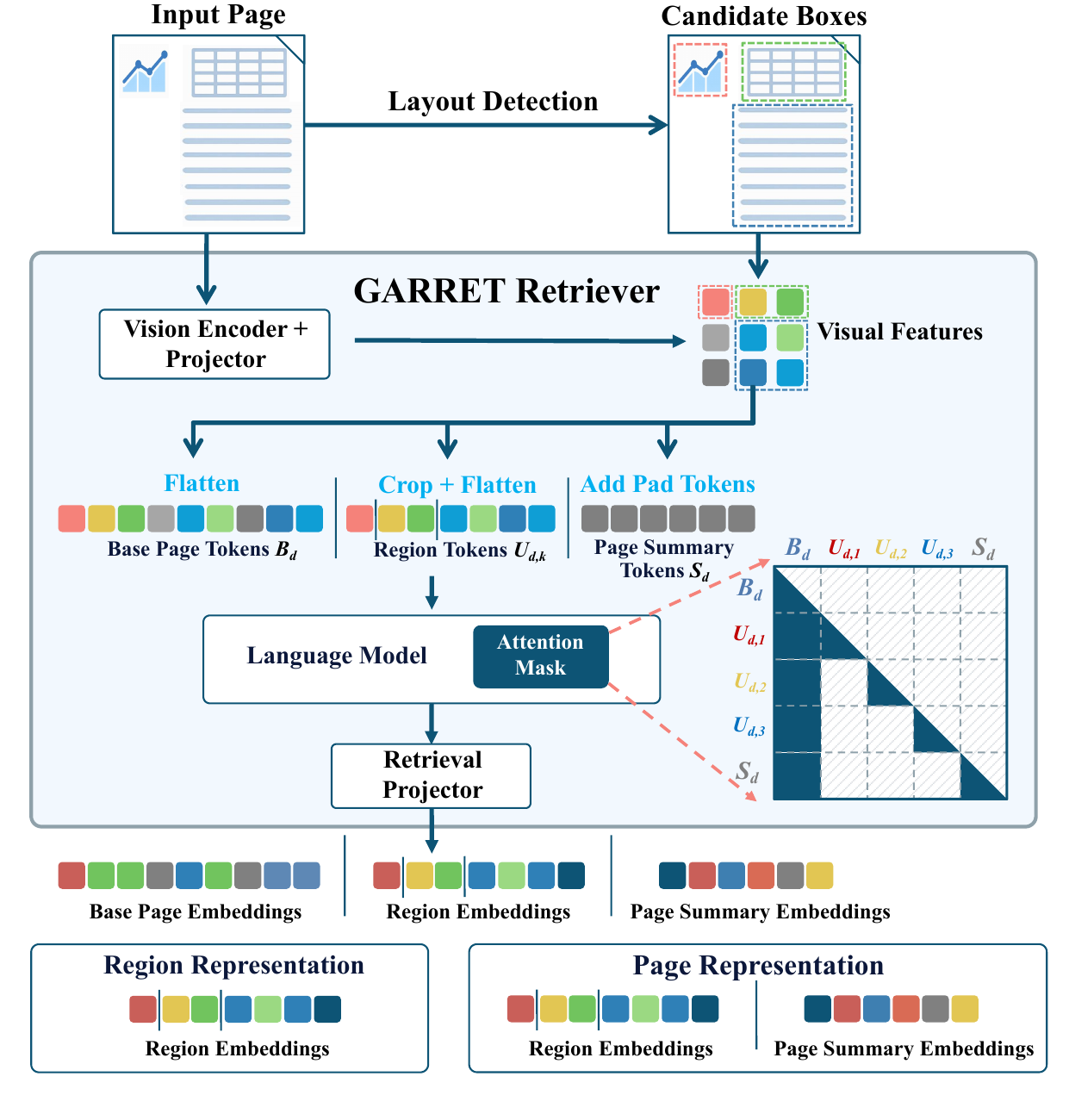}
    \caption{Architecture of GARRET's context-preserving multi-granularity retriever. A specialized attention mask allows region tokens to absorb global page context without mutual interference. After projection, base page embeddings are discarded. The retained region and summary embeddings form a highly compact representation, enabling both fine-grained region retrieval and efficient page-level retrieval.}
    \label{fig:retriever}
\end{figure}

\section{Related Work}

\subsection{Fine-Grained Visual Document Retrieval}

Vision-native document retrievers encode rendered pages to preserve layout and
visual content. DSE and VisRAG use one vector per page, whereas ColPali applies
ColBERT-style late interaction
\citep{ma2024dse,yu2024visrag,khattab2020colbert,faysse2024colpali}.
Follow-up work analyzes late interaction, compresses multi-vector indexes, or
forms page-contextualized visual chunks
\citep{qiao2025reproducibilityvdr,ma2025storageefficientvdr,
yan2026visuallatechunking}. These systems retain the page as the retrieval
output.

RegionRAG forms query-dependent regions at inference, whereas LFRAG indexes
predefined layout blocks with explicit query--block labels
\citep{li2026regionrag,zhu2026lfraglayoutorientedfinegrainedretrievalaugmented}.
AGREE and ReAlign use MLLM attention, localized regions, or generated
descriptions to improve page retrieval
\citep{cui2025attentiongroundedvdr,yang2026realign}. Spatially-Grounded
Retrieval propagates page-patch scores to OCR regions, ColParse combines
parsed subimages with a global page vector, and VISA generates answers with
supporting boxes
\citep{georgiou2025spatialretrieval,yan2026beyondgrid,ma2025visa}. These
methods do not learn persistent evidence units directly from QA supervision.

Training region retrievers therefore remains dependent on costly manual
query--region annotations or large-model-generated supervision, which is
computationally expensive and may introduce noisy fixed targets. GARRET instead
learns latent query--region relevance from
$(\text{page},\text{question},\text{answer})$ triplets without explicit region
labels. Existing benchmarks likewise focus primarily on page ranking. Although
ViDoRe V3 provides evidence boxes, they support localization within a relevant
page rather than corpus-wide region retrieval
\citep{loison2026vidorev3}. We align these annotations with query-independent
layout boxes to derive fixed candidate regions and region-level relevance
judgments for corpus-wide evaluation.

\subsection{Generative Supervision for Retrieval}

ReVela connects language modeling with dense retrieval by optimizing textual
document representations through a language-modeling objective, while ElicitR
combines contrastive retrieval learning with a generative regularizer
\citep{cai2026revela,cai2026elicitr}. These methods demonstrate that generative
objectives can provide supervision beyond explicit relevance labels, but they
operate on textual passages, whose retrieval units are naturally defined and
share the same modality as the generator.

GARRET extends this principle to multimodal region retrieval. Visual regions
must be spatially defined, preserve their parent-page context, and remain
independently indexable. GARRET treats their relevance as latent and uses soft
region routing to connect the retriever with an auxiliary VLM. By jointly
predicting the paired question and answer, the generation objective supervises
both query--region alignment and evidence sufficiency. Its gradients directly
update the persistent region embeddings retained in the retrieval index,
without teacher-generated attention labels, descriptions, or query--region
annotations.

\section{Method}
\label{sec:method}

Figure~\ref{fig:method} gives an overview of \textbf{GARRET}, a fine-grained visual
document retriever trained using
$(\text{page}, \text{question}, \text{answer})$ triplets without
query-to-region annotations. A shared context-preserving retriever
produces both page embeddings and region embeddings. During training, an
auxiliary VLM backpropagates the question--answer generation loss through the
retrieval scores to learn latent region relevance. The auxiliary VLM is
discarded at inference.

\subsection{Problem Formulation}

Let $\mathcal{D}$ be a corpus of visual document pages. For each page $d$,
MinerU \citep{wang2024mineru} extracts $m_d$ query-independent layout candidates
$\mathcal{R}(d)=\{r_{d,k}\}_{k=1}^{m_d}$, where each candidate $r_{d,k}$ is defined by a
bounding box $b_{d,k}$ and may correspond to a text block, table, figure, or
other layout element.

Training uses $\mathcal{T}=\{(d_i^+,q_i,a_i)\}_{i=1}^{N}$, where the paired
page $d_i^+$ contains evidence for answer $a_i$ to question $q_i$, but its
supporting regions are unannotated. We therefore treat region
relevance as a latent variable. The retriever learns a page-level score
$s_D(q,d)$ and a region-level score $s_R(q,d,k)$, which are combined to rank
bounding-box regions across the corpus.

\subsection{Context-Preserving Multi-Granularity Retriever}
\label{sec:multi_granularity}

To efficiently support both region- and page-level retrieval, GARRET derives
contextual region embeddings and a compact page representation through a
shared page encoding, as shown in Figure~\ref{fig:retriever}.

Given a page image $d$, the vision encoder and visual projector produce a
spatial feature map $M_d$. We flatten the complete feature map into the
\textbf{base page tokens} $B_d$. For each layout candidate $r_{d,k}$ with
bounding box $b_{d,k}$, we crop the corresponding area from $M_d$ and
flatten it into a region token group:
\[
\begin{gathered}
B_d=\operatorname{Flatten}(M_d),\\
U_{d,k}
=
\operatorname{Flatten}
\left(
\operatorname{Crop}(M_d,b_{d,k})
\right).
\end{gathered}
\]
The number of tokens in $U_{d,k}$ depends on the spatial extent of the
candidate on the feature map. Since all region groups are extracted from
the same $M_d$, each page requires only one vision-encoder forward pass.

We concatenate the base page tokens, all region token groups, and $n_s$
\textbf{page summary tokens} $S_d^{(0)}$:
\[
X_d
=
B_d
\mathbin\Vert U_{d,1}
\mathbin\Vert\cdots\mathbin\Vert U_{d,m_d}
\mathbin\Vert S_d^{(0)}.
\]
We use $n_s=100$ summary tokens in all experiments and initialize them with
padding-token embeddings, following ColPali \citep{faysse2024colpali}. These tokens provide an additional global readout
for page content that may not be fully covered by the layout candidates.

A block-structured causal attention mask controls the information flow.
Base page tokens follow standard causal attention and cannot access the
appended tokens. Each region token attends to all base page tokens and
preceding tokens within its own region group, but not to tokens from other
regions. Summary tokens similarly attend to all base page tokens and
preceding summary tokens, while remaining isolated from the region groups.
Therefore, each region incorporates global page information without being
contaminated by other candidate regions. This design preserves the
surrounding context needed to interpret local content while keeping every
region independently retrievable, forming a visual analogue of late
chunking \citep{gunther2024latechunking,yan2026visuallatechunking}.

Let $H_{d,k}^{R}$ and $H_d^{S}$ denote the contextualized hidden states at
the $k$-th region group and the summary-token positions, respectively.
After applying the retrieval projector $P_\theta$ and token-wise
$\ell_2$ normalization, the region representation is
\[
T_R(d,k)
=
\operatorname{norm}
\left(
P_\theta(H_{d,k}^{R})
\right),
\]
and the page representation is
\[
T_D(d)
=
\operatorname{norm}
\left(
P_\theta
\left(
H_{d,1}^{R}
\mathbin\Vert\cdots\mathbin\Vert
H_{d,m_d}^{R}
\mathbin\Vert H_d^{S}
\right)
\right).
\]
Thus, each contextualized region group serves two retrieval granularities:
$T_R(d,k)$ is indexed as an independent evidence unit, while the same group
is reused as part of the parent-page representation $T_D(d)$. The base page
hidden states are used only to provide context during offline encoding and
are discarded before indexing. GARRET therefore stores only the region
embeddings and the small set of page-summary embeddings.

Although appending region and summary tokens lengthens document-side encoding,
this cost is incurred only once during offline index construction and does not
affect online retrieval \citep{cui2025thinkthenembed}. Since document pages
typically contain substantial whitespace, the retained region-token groups
cover far fewer positions than the dense full-page grid, resulting in a
shorter indexed representation.

Following late-interaction visual retrieval
\citep{faysse2024colpali}, the query encoder maps $q$ to
$Q(q)=\{\mathbf q_u\}_{u=1}^{n_q}$. For a document-side multivector
$T=\{\mathbf t_v\}_{v=1}^{n_T}$, we use query-length-normalized MaxSim
\citep{khattab2020colbert}:
\[
s(q,T)
=
\frac{1}{n_q}
\sum_{u=1}^{n_q}
\max_{1\leq v\leq n_T}
\left\langle
\mathbf q_u,\mathbf t_v
\right\rangle.
\]
The page- and region-level scores are
\[
\begin{gathered}
s_D(q,d)=s(q,T_D(d)),\\
s_R(q,d,k)=s(q,T_R(d,k)).
\end{gathered}
\]
Finally, we combine local region relevance with the relevance of its parent
page:
\[
F(q,d,k)=s_R(q,d,k)+s_D(q,d).
\]
The region term identifies fine-grained evidence, while the page term
disambiguates similar local regions using their global document context.
The same score is used for training-time routing and inference-time region
ranking.

\subsection{Generative Feedback for Region Retrieval}
\label{sec:generative_feedback}

\paragraph{Differentiable region routing.}
As illustrated in Figure~\ref{fig:method}, GARRET converts the retriever's
page and region scores into routing weights that directly control the visual
evidence available to an auxiliary VLM. This design places region selection
inside the generation path, allowing the next-token prediction (NTP) loss to
supervise the retriever without query-to-region annotations. For training
example $i$, let $\mathcal{C}_i$ denote the candidate region pool defined in the Candidate Sampling subsection. For a candidate $r=(d,k)$, we use the same
combined score as region retrieval,
$F_{i,r}=F(q_i,d,k)=s_R(q_i,d,k)+s_D(q_i,d)$, and normalize the scores into
a differentiable routing distribution:
\[
\pi_{i,r}
=
\frac{\exp(F_{i,r}/\tau_r)}
{\sum_{r'\in\mathcal{C}_i}\exp(F_{i,r'}/\tau_r)},
\qquad r\in\mathcal{C}_i,
\]
where $\tau_r$ controls the routing sharpness. Thus, $\pi_{i,r}$ represents
the retriever's current belief that region $r$ provides useful evidence for
query $q_i$.

\paragraph{Isolated visual sources.}
For each candidate $r=(d,k)$, we crop the pixels within its layout bounding
box, $C_r=\operatorname{Crop}(d,b_{d,k})$, and form
$\mathcal{I}_i=\{C_r:r\in\mathcal{C}_i\}$. Each crop is encoded independently
by the auxiliary VLM and treated as a separate visual source. This isolation
is essential: the contribution controlled by $\pi_{i,r}$ can contain only
evidence inside the corresponding bounding box, preventing neighboring
regions from leaking information into the generation objective. The crop
features are used only to provide training-time feedback and are not stored
in the retrieval index.

\paragraph{Retriever-weighted in-batch cross-attention.}
We insert an in-batch cross-attention branch into selected decoder layers of
the auxiliary VLM. Candidate crop encodings in the local pool serve as source
memories, while the paired question--answer sequence forms the target branch.
At decoder layer $\ell$, the source branch provides stop-gradient visual keys
and values $(K_r^\ell,V_r^\ell)$ for each candidate. For target token $t$,
the query $Q_{i,t}^\ell$ first attends to each crop independently and then
aggregates the resulting crop-specific contexts using the retriever-derived
routing weights:
\begin{align*}
b_{i,t,r}^{\ell}
&=
\operatorname{softmax}
\left(
\frac{Q_{i,t}^{\ell}(K_r^\ell)^\top}{\sqrt{d_h}}
\right)V_r^\ell,\\
b_{i,t}^{\ell}
&=
\sum_{r\in\mathcal{C}_i}
\pi_{i,r}\,b_{i,t,r}^{\ell},\\
h_{i,t}^{\ell}
&=
s_{i,t}^{\ell}+b_{i,t}^{\ell},
\end{align*}
where $d_h$ is the attention-head dimension and $s_{i,t}^{\ell}$ denotes the
ordinary causal self-attention output. The first operation performs
token-level selection within each crop, whereas the weighted sum performs
region-level routing across candidates. Keeping the crop-wise attention
softmaxes separate ensures that $\pi_{i,r}$ acts as an explicit region gate
rather than being entangled with token competition across regions. This
construction establishes the gradient path
$\mathcal{L}_{\mathrm{NTP}}\!\rightarrow\!\pi_{i,r}\!\rightarrow\!F_{i,r}
\!\rightarrow\!\{s_R,s_D\}\!\rightarrow\!\theta$: regions that reduce the
generation loss receive positive retrieval feedback, while distractors are
down-weighted.

\paragraph{Joint question and answer generation.}
Given a fixed instruction, the auxiliary VLM is trained with teacher forcing
to generate
\[
y_i=
\bigl[
\texttt{Question:},\,q_i,\,
\texttt{Answer:},\,a_i
\bigr].
\]
The two targets provide complementary supervision. Predicting the question
encourages the routed regions to align with the query semantics, whereas
predicting the answer requires them to contain sufficient evidence for the
correct response. Their combination also limits target-specific shortcuts:
answer-only training may exploit language priors for short answers, while
question-only training may favor topically related but non-evidential
regions. The NTP objective is
\[
\begin{aligned}
\mathcal{L}_{\mathrm{NTP}}
&= -\frac{1}{|\mathcal{B}|}\sum_{i\in\mathcal{B}}\frac{1}{L_i}
\\[-2pt]
&\quad\times\sum_{t=1}^{L_i}\log p_{\phi,\theta}
\left(y_{i,t}\mid y_{i,<t},\mathcal{I}_i,\boldsymbol{\pi}_i\right).
\end{aligned}
\]
where $L_i$ is the target length, $\phi$ denotes the auxiliary VLM
parameters, and $\theta$ denotes the retriever parameters. Unlike
pseudo-labeling, GARRET does not convert VLM outputs into fixed region
labels. Instead, the NTP loss provides continuous, query-conditioned
generative feedback to the page and region representations that are later
used for retrieval.

\subsection{Joint Optimization and Inference}
\label{sec:joint_training}

\paragraph{Candidate sampling.}
Computing cross-attention over every in-batch region would exceed practical
memory limits. We therefore construct a compact candidate pool from the paired
page, hard-negative pages, and randomly sampled negative pages.

Hard negatives are the top-$H$ nonpaired pages according to their
highest-scoring region:
\[
\begin{aligned}
\mathcal{H}_i
&=
\operatorname{TopH}_{d\neq d_i^+}
\left[
\max_{1\leq k\leq m_d}F(q_i,d,k)
\right],\\
\mathcal{P}_i
&=
\{d_i^+\}
\cup\mathcal{H}_i
\cup\mathcal{N}_i^{\mathrm{rand}},\\
\mathcal{C}_i
&=
\{(d,k):d\in\mathcal{P}_i,\;1\leq k\leq m_d\},
\end{aligned}
\]
where $\mathcal{N}_i^{\mathrm{rand}}$ contains randomly sampled nonpaired pages
outside $\mathcal{H}_i$. Hard negatives provide visually or semantically
confusable candidates, while random negatives broaden coverage. All regions
from the selected pages remain candidates, including distractor regions on the
paired page.

\paragraph{Optimization and deployment.}
The NTP loss learns latent region relevance, while a ColPali-style pairwise
softplus loss with in-batch hard negatives preserves page-level retrieval
quality \citep{faysse2024colpali}. The joint objective is
\[
\mathcal{L}
=
\mathcal{L}_{\mathrm{page}}
+
\lambda_{\mathrm{NTP}}\mathcal{L}_{\mathrm{NTP}}.
\]
We optimize the retriever's language-model and projection adapters and the
auxiliary VLM's language-model adapter, while keeping both vision encoders
frozen. At inference, the auxiliary VLM and crop-source branch are discarded,
leaving the retriever to rank pages by $s_D(q,d)$ and all candidate regions by
$F(q,d,k)$.

\subsection{Region-Level Benchmark Construction}

Existing visual document retrieval benchmarks mainly evaluate page-level ranking. Although ViDoRe V3 provides human-annotated evidence boxes, these free-form annotations are intended for localization within relevant pages rather than corpus-wide region retrieval \citep{loison2026vidorev3}. We retain the complete query set and eight domains of ViDoRe V3, and use MinerU \citep{wang2024mineru} to extract query-independent layout bounding boxes from every page. Layout boxes that overlap an original evidence box, i.e., with $\operatorname{IoU}>0$, are selected as candidate annotations. This procedure produces a fixed corpus of 200,682 layout regions, which are indexed and ranked directly during evaluation. This setup is consistent with our formulation of MinerU regions as persistent retrieval units rather than query-dependent crops.

For each overlapping layout box, we draw the box on the full page and ask Qwen3-VL-32B \citep{bai2025qwen3vl} to score its relevance to the query from 0 to 5. Regions with scores greater than 4 are initially labeled positive. Two annotators then independently review all candidate labels. A region is considered relevant only when the content inside the box provides direct or indispensable evidence for answering the query; regions with only topical similarity, incidental geometric overlap, or evidence located primarily outside the box are rejected. Disagreements are discussed until consensus is reached. The final human-verified binary labels are used to evaluate corpus-wide region ranking with nDCG.

\begin{table*}[t]
    \centering
    \small
    \renewcommand{\arraystretch}{1.14}
    \begin{tabular*}{0.96\textwidth}{@{\extracolsep{\fill}}l*{10}{r}@{}}
        \toprule
        \textbf{Method} & \textbf{Avg.} & \textbf{CS} & \textbf{Ene.}
        & \textbf{Fin-EN} & \textbf{Fin-FR} & \textbf{HR} & \textbf{Ind.}
        & \textbf{Pha.} & \textbf{Phy.} & \textbf{Vec./reg.}\\
        \midrule
        \multicolumn{11}{l}{\textit{Text-based Document Retrievers}} \\
        BM25 & 16.24 & 25.93 & 5.97 & 19.67 & 5.47 & 28.70 & 18.28 & 21.84 & 4.05 & -- \\
        BGE-M3 & 25.26 & 33.32 & 29.63 & 26.17 & 12.57 & 35.51 & 24.45 & 22.17 & 18.28 & 1.0 \\
        Qwen3-Emb.-8B & 35.22 & 49.97 & \underline{38.17} & \underline{35.09} & 19.20 & 38.67 & 31.72 & 33.20 & 35.77 & 1.0 \\
        \midrule
        \multicolumn{11}{l}{\textit{Vision-based Document Retrievers}} \\
        ColPali-v1.3 & 31.51 & 50.15 & 28.51 & 27.98 & 14.26 & 39.74 & 31.18 & 30.47 & 29.75 & 1031.0 \\
        ColQwen2.5-v0.2 & \underline{38.32} & \textbf{56.46} & 37.70 & \textbf{35.30} & \underline{22.07} & \underline{45.36} & \underline{37.40} & 35.88 & \underline{36.41} & 160.1 \\
        RegionRAG & 33.77 & 54.61 & 32.35 & 26.22 & 16.05 & 37.97 & 31.06 & \underline{41.29} & 30.63 & 29.4 \\
        \textbf{GARRET} & \textbf{41.26} & \underline{55.46} & \textbf{42.47} & 34.05 & \textbf{22.37} & \textbf{46.07} & \textbf{41.93} & \textbf{49.72} & \textbf{38.04} & 28.3 \\
        \bottomrule
    \end{tabular*}
    \caption{Region retrieval on \textsc{ViDoRe-V3-Region} (nDCG@10, $\times 100$). GARRET achieves the best average performance with the fewest indexed vectors among dense visual retrievers. Bold and underlined values denote the best and second-best results.}
    \label{tab:v3_main}
\end{table*}

\begin{table}[t]
    \centering
    \footnotesize
    \setlength{\tabcolsep}{1.4pt}
    \renewcommand{\arraystretch}{1.14}
    \begin{tabular*}{\columnwidth}{@{\extracolsep{\fill}}lrrrrr@{}}
        \toprule
        \textbf{Method} & \multicolumn{2}{c}{\textbf{nDCG}}
        & \multicolumn{3}{c}{\textbf{Layout Recall}} \\
        \cmidrule(lr){2-3}\cmidrule(l){4-6}
        & \textbf{@5} & \textbf{@10}
        & \textbf{R@1} & \textbf{R@5} & \textbf{R@10} \\
        \midrule
        \multicolumn{6}{l}{\textit{Text-based regions}} \\
        BM25-OCR        & 32.73 & 34.65 & 20.82 & 36.00 & 42.17 \\
        BM25-VLM        & 41.68 & 44.21 & 28.72 & 50.78 & 58.70 \\
        BGE-M3          & 42.48 & 45.58 & 27.90 & 50.88 & 59.75 \\
        Qwen3-Emb.-8B   & 42.14 & 45.32 & 24.87 & 47.98 & 56.73 \\
        \addlinespace[2pt]
        \multicolumn{6}{l}{\textit{Independent visual crops}} \\
        ColPali-v1.3    & \underline{51.31} & \underline{54.83}
                        & \underline{36.95} & 59.02 & 67.75 \\
        ColQwen2.5-v0.2 & 50.90 & 54.08 & 36.72
                        & \underline{59.90} & \underline{68.40} \\
        RegionRAG       & 50.03 & 53.14 & 35.58 & 58.87 & 68.26 \\
        \addlinespace[2pt]
        \multicolumn{6}{l}{\textit{Page-context region encoding}} \\
        \textbf{GARRET} & \textbf{56.68} & \textbf{60.04}
                        & \textbf{41.30} & \textbf{64.07} & \textbf{73.14} \\
        \bottomrule
    \end{tabular*}
    \caption{Region retrieval on MMDocIR. All metrics are reported on a $\times 100$ scale. GARRET achieves the best nDCG and layout recall at all evaluated cutoffs, demonstrating strong cross-benchmark generalization. Bold and underlined values denote the best and second-best results.}
    \label{tab:mmdocir_main}
\end{table}

\section{Experiments}
\label{sec:experiments}

\subsection{Experimental Setup}

\paragraph{Training Details.}
To ensure a fair comparison, we train GARRET exclusively on the official ViDoRe training set (118,195 triplets) used by the ColPali and ColQwen series, without relying on external or synthetic data. MinerU \cite{wang2024mineru} is solely used for layout parsing. Both the retriever and the auxiliary VLM are initialized from Qwen2.5-VL-3B-Instruct \cite{bai2025qwen25vl} with their vision encoders frozen. The retriever uses LoRA \cite{hu2022lora} (rank 64, $\alpha=128$) and produces 128-dimensional embeddings, while the VLM uses LoRA (rank 32, $\alpha=64$). We train for 5 epochs on 16 NVIDIA A100 (80GB) GPUs with a learning rate of $1\times10^{-4}$, and a per-GPU batch size of 24. The number of page summary tokens is set to 100. The NTP loss weight is set to 0.1, and the routing temperature to 0.01.

\paragraph{Benchmarks and Metrics.}
We evaluate across three main dimensions: region retrieval, page retrieval, and downstream generation. For region-level retrieval, we evaluate on our proposed \textsc{ViDoRe-V3-Region} (2,419 queries, 200,682 regions) using nDCG@10 as the primary metric. To assess cross-benchmark generalization, we additionally evaluate on MMDocIR \cite{dong2025mmdocir} for in-document region localization. For page-level retrieval, we follow the official protocols of ViDoRe V1, V2, and V3 \cite{faysse2024colpali,mace2025vidorev2,loison2026vidorev3}. Furthermore, we report the number of indexed vectors per page/region to compare storage efficiency.

\paragraph{Baselines.}
We compare GARRET against lexical models like BM25 \cite{robertson2009bm25}; text-dense models including BGE-M3 \cite{chen2024m3embedding} and Qwen3-Embedding-8B \cite{zhang2025qwen3embedding}; and visual retrievers such as ColPali, ColQwen \cite{faysse2024colpali}, and RegionRAG \cite{li2026regionrag}. For text-based baselines, we encode parsed text. Since ColPali and ColQwen are inherently page-level retrievers, we adapt them for region retrieval by cropping and encoding each benchmark bounding box as an independent image. For RegionRAG, we map its dynamically predicted regions to our benchmark boxes. As LFRAG \cite{zhu2026lfraglayoutorientedfinegrainedretrievalaugmented} is not open-source, we evaluate its core architectural design via controlled ablations.

\paragraph{Downstream RAG Evaluation.}
To measure the practical impact of fine-grained retrieval, we conduct a RAG experiment on the ViDoRe V3 test set using Qwen2.5-VL-7B-Instruct as the generator. Page-level baselines are provided with the top-5 retrieved full pages as context. In contrast, GARRET is provided with the top-$k$ retrieved region crops ($k \in \{5, 10, 15, 20\}$). We use Qwen2.5-VL-32B-Instruct as an automated judge to evaluate strict answer accuracy, reporting the percentage of fully correct answers.

\subsection{Region-Level Retrieval Performance}

We first evaluate GARRET on our newly constructed \textsc{ViDoRe-V3-Region} benchmark, with results presented in Table~\ref{tab:v3_main}. GARRET establishes a new state of the art with a macro-average nDCG@10 of 41.26, substantially outperforming the strongest vision-based baseline by 2.94 points (Table~\ref{tab:v3_main}). This superiority stems from learning persistent, page-contextualized region embeddings via generative feedback. Unlike baselines that either lose context by encoding isolated crops (ColQwen) or use unstable dynamic regions (RegionRAG), GARRET creates robust representations without costly annotations. This corresponds to 82.3\% fewer vectors per region than the ColQwen adaptation and a substantially shorter indexed representation.

To assess generalization, we evaluate GARRET on the MMDocIR benchmark, which focuses on in-document region localization. As shown in Table~\ref{tab:mmdocir_main}, GARRET demonstrates strong transferability, outperforming all baselines across both nDCG and Layout Recall metrics. It achieves an nDCG@10 of 60.04, surpassing the next-best method (ColPali-v1.3) by 5.21 points. This consistent lead on a different benchmark underscores the robustness of our context-aware region representations.

\subsection{Page-Level Retrieval Performance}

A key advantage of our multi-granularity architecture is its ability to enhance region retrieval without compromising page-level performance. As shown in Table~\ref{tab:page_retrieval}, GARRET not only preserves but often matches or exceeds the performance of leading page-level retrievers. Specifically, GARRET achieves state-of-the-art nDCG@10 scores on ViDoRe V1 (88.56) and V3 (57.29), slightly outperforming ColQwen2.5-v0.2, and remains highly competitive on V2.

Crucially, GARRET achieves this competitive performance while being significantly more efficient. GARRET indexes only 392.9 vectors per page, representing a 47.8\% reduction compared to ColQwen2.5 and RegionRAG, and a 61.9\% reduction compared to ColPali. This result validates our design: the context-preserving multi-granularity embedding paradigm successfully creates a compact, shared index that excels at fine-grained retrieval without sacrificing coarse-grained accuracy or incurring a storage penalty.

\begin{table}[t]
    \centering
    \footnotesize
    \setlength{\tabcolsep}{4.0pt}
    \renewcommand{\arraystretch}{1.14}
    \begin{tabular*}{\columnwidth}{
        @{\extracolsep{\fill}}lrrrr@{}
    }
        \toprule
        \textbf{Method}
        & \textbf{V1}
        & \textbf{V2}
        & \textbf{V3}
        & \textbf{Vec./page}\\
        \midrule

        \multicolumn{5}{l}{\textit{Text-based Document Retrievers}} \\
        BM25
            & 69.94
            & 21.61
            & 38.29
            & - \\
        BGE-M3
            & 68.27
            & 49.75
            & 44.63
            & 1.0 \\
        Qwen3-Emb.-8B
            & 75.82
            & 55.28
            & 53.93
            & 1.0 \\

        \midrule
        \multicolumn{5}{l}{\textit{Vision-based Document Retrievers}} \\
        ColPali-v1.3
            & 85.42
            & 57.81
            & 48.86
            & 1031.0 \\
        ColQwen2.5-v0.2
            & \underline{88.45}
            & \underline{58.59}
            & \underline{57.18}
            & 752.3 \\
        RegionRAG
            & 86.70
            & \textbf{59.85}
            & 57.14
            & 752.3 \\
        \textbf{GARRET}
            & \textbf{88.56}
            & 57.51
            & \textbf{57.29}
            & 392.9 \\

        \bottomrule
    \end{tabular*}
    \caption{Page retrieval on ViDoRe V1--V3 using nDCG@10 ($\times 100$). GARRET remains competitive with the strongest visual retrievers while using substantially fewer indexed vectors per page. Bold and underlined values denote the best and second-best results.}
    \label{tab:page_retrieval}
\end{table}

\subsection{Impact on Downstream RAG Evaluation}

To evaluate the practical utility of fine-grained retrieval, we measure its impact on downstream RAG performance on ViDoRe V3, with results in Table~\ref{tab:rag_area}. We compare standard page-level retrieval, where the top-5 full pages are fed to the generator, against our region-level retrieval, where the top-\(k\) regions from GARRET are used.

The results demonstrate that retrieving precise regions is more effective and efficient than retrieving entire pages. The best page-level baseline, ColQwen2.5-v0.2, achieves a strict accuracy of 37.00\% using five full pages as context. In contrast, GARRET surpasses this performance by retrieving only 15 regions, reaching 38.00\% accuracy while using just 49.51\% of the context area (\textit{Area/P5}). By increasing the number of retrieved regions to 20, GARRET's accuracy further improves to 39.12\%, using only 66.08\% of the page-level context area. This shows that providing the generator with a smaller set of more relevant, evidence-rich regions leads to higher-quality answers and reduces the processing of irrelevant information, highlighting the tangible benefits of fine-grained retrieval for RAG systems.

\begin{table}[t]
    \centering
    \footnotesize
    \setlength{\tabcolsep}{1.4pt}
    \renewcommand{\arraystretch}{1.14}
    \begin{tabular*}{\columnwidth}{@{\extracolsep{\fill}}llrr@{}}
        \toprule
        \textbf{Method / context} & \textbf{Units} & \textbf{Area/P5}
        & \textbf{Acc} \\
        \midrule
        \multicolumn{4}{l}{\textit{Page-level retrieval}} \\
        BM25             & 5 pages & 100\% & 27.53 \\
        BGE-M3           & 5 pages & 100\% & 33.85 \\
        Qwen3-Emb.-8B    & 5 pages & 100\% & 36.71 \\
        ColPali-v1.3     & 5 pages & 100\% & 33.51 \\
        ColQwen2.5-v0.2  & 5 pages & 100\% & 37.00 \\
        \midrule
        \multicolumn{4}{l}{\textit{Region-level retrieval}} \\
        \multirow{4}{*}{GARRET}
            & 5 regions  & 16.36\% & 35.77 \\
            & 10 regions & 32.93\% & 36.89 \\
            & 15 regions & 49.51\% & \underline{38.00} \\
            & 20 regions & 66.08\% & \textbf{39.12} \\
        \bottomrule
    \end{tabular*}
    \caption{RAG results on ViDoRe V3. Page-level methods use the top-5
retrieved pages, while GARRET uses the top-\(k\) retrieved layout regions.
Area/P5 denotes retrieved area relative to five full pages. Acc is reported
as a percentage.}
    \label{tab:rag_area}
\end{table}

\begin{table}[t]
    \centering
    \footnotesize
    \setlength{\tabcolsep}{2.5pt}
    \renewcommand{\arraystretch}{1.14}
    \begin{tabular*}{\columnwidth}{@{\extracolsep{\fill}}lrrr@{}}
        \toprule
        \textbf{Variant} & \textbf{Region} & \textbf{Page}
        & \textbf{Vectors/page} \\
        \midrule

        \textbf{GARRET}
        & \textbf{41.26}
        & \textbf{57.29}
        & 392.9 \\

        \midrule
        \multicolumn{4}{l}{\textit{Generative-target controls}} \\
        w/o generative feedback
        & 36.33
        & \underline{57.04}
        & 392.9 \\
        Answer target only
        & \underline{41.02}
        & 56.92
        & 392.9 \\
        Question target only
        & 39.98
        & 56.89
        & 392.9 \\

        \midrule
        \multicolumn{4}{l}{\textit{Model architecture variants}} \\
        w/o page summary tokens
        & 40.06
        & 55.82
        & 292.9 \\
        LFRAG-style structure
        & 26.58
        & 53.68
        & 1408.3 \\
        RegionRAG-style structure
        & 32.46
        & 45.55
        & 752.3 \\

        \bottomrule
    \end{tabular*}
    \caption{Ablation studies on ViDoRe V3 using nDCG@10 ($\times 100$). Generative feedback primarily improves region retrieval, while joint question--answer generation and the full shared architecture yield the best performance. Bold and underlined values denote the best and second-best results.}
    \label{tab:garret_ablation}
\end{table}

\subsection{Ablation Studies}

We conduct controlled ablations on ViDoRe V3
(Table~\ref{tab:garret_ablation}). Each variant modifies only the component
indicated by its name; all non-ablated components, training data, candidate
boxes, initialization, sampling strategy, optimization settings, training
budget, and evaluation protocol remain fixed. In particular, the LFRAG-
and RegionRAG-inspired variants replace only the representation structure
while retaining GARRET's NTP feedback, page-level loss, and routing
mechanism. They therefore serve as internal architectural controls rather
than reproductions of the original systems.

\paragraph{Generative-Target Controls.}
Removing generative feedback, including the auxiliary VLM and NTP
objective, reduces region nDCG@10 from 41.26 to 36.33, while page
performance changes only from 57.29 to 57.04. This confirms that the
page-level objective largely preserves document relevance, whereas
generative feedback supplies the fine-grained signal needed to distinguish
evidence-bearing regions. When only the generative target is changed,
answer-only training reaches 41.02 region nDCG@10, outperforming
question-only training at 39.98. Joint question--answer generation performs
best, supporting the complementary roles of query--region semantic
alignment and answer-evidence sufficiency.

\paragraph{Model Architecture Variants.}
Removing page summary tokens lowers both region and page retrieval
performance, indicating that they provide useful global information beyond
the retained region groups. Under the same training protocol, the
LFRAG-inspired independent-encoding structure and the RegionRAG-inspired
structure also underperform GARRET. Differences in vectors per page are
direct consequences of the corresponding representation design rather than
additional experimental changes. These results support the value of
GARRET's shared, page-contextualized representations and persistent layout
regions, which jointly improve retrieval quality while maintaining a
compact index.

\section{Conclusion}

We introduced GARRET, a novel framework for fine-grained region retrieval in visual documents that does not require manual query-to-region training annotations. GARRET's core innovation is learning region relevance from generative feedback, where the next-token prediction loss from an auxiliary VLM provides a powerful, annotation-free training signal. This is realized through a retriever-weighted cross-attention mechanism. Furthermore, our context-preserving multi-granularity embedding architecture, a visual analogue to late chunking, creates a unified, layout-aware index for both page and region retrieval, achieving state-of-the-art performance with significantly fewer indexed vectors.

On our new large-scale benchmark, GARRET establishes a new state of the art with an nDCG@10 of 0.4126. By enabling the training of high-performance, efficient region retrievers without manual labels, GARRET represents a significant step toward building more scalable, accurate, and cost-effective multimodal information systems.

\bibliography{references}
\par\bigskip

\clearpage
\appendix

\section{Implementation Details}
\label{sec:supp_implementation}

This section specifies the implementation choices omitted from the main paper.
We first describe the construction and encoding of persistent layout candidates,
and then detail the training-only generative-feedback path.  In particular, the
auxiliary VLM applies retriever-weighted cross-attention in \emph{all} of its
language layers.  The auxiliary branch is removed after training and therefore
does not affect the deployed index or online query encoding.

Table~\ref{tab:supp_config} summarizes the principal model and optimization
settings.  GARRET is trained on 118,195 page--question--answer triplets from the
official ViDoRe training set.  We use neither additional external data nor
newly synthesized examples, and no query-to-region relevance labels are
provided.

\subsection{Training Data and Candidate Preparation}

\paragraph{Training Records.}
Each training record contains a question $q_i$, its answer $a_i$, one paired
page $d_i^+$, and the query-independent layout boxes extracted from that page.
The paired page is the only page-level supervision supplied to the retriever.
Before batching, we reject records with an empty question, an invalid answer,
no valid layout candidate, or anything other than exactly one paired page.
When an answer contains multiple acceptable strings, we remove empty entries
and join the remaining strings with semicolons.  For multiple-choice records
whose answer is a letter, we replace the letter with the corresponding option
text when that mapping is available.  These operations only canonicalize the
generation target and never create region labels.

\paragraph{Supervision Boundary.}
The data preprocessing artifacts may contain legacy positive- and
negative-box fields used by other training variants.  GARRET explicitly clears
these fields before the document forward pass and constructs its region pool
exclusively from the full candidate list.  Consequently, neither the region
embeddings nor the routing distribution can access query-to-box labels.  The
only query-conditioned supervision for region selection is the gradient of the
question--answer generation loss.

\paragraph{Candidate Canonicalization.}
MinerU provides a fixed ordered list of layout boxes for
each page.  We express every box as $(x_1,y_1,x_2,y_2)$ in the normalized
$[0,1000]^2$ page coordinate system, clamp its coordinates to the page
boundary, and remove entries with non-finite coordinates or non-positive width
or height.  Valid boxes retain their original relative order.  The same
canonicalization routine is used by the retrieval-side feature cropper and the
feedback-side pixel cropper.  Each surviving box is assigned a group index;
the tuple consisting of its page index and group index identifies the
corresponding region tokens, pixel crop, routing weight, and retrieval result.
Training aborts if these two candidate views differ in count or order.

\subsection{Context-Preserving Retriever Implementation}

\paragraph{Page and Region Inputs.}
The retriever processes each rendered page once, with at most 768 base visual
tokens.  Let $M_d$ denote the spatial feature map produced by the vision
encoder and visual projector.  For each canonical box $b_{d,k}$, we map its
normalized coordinates to $M_d$, crop the corresponding feature-grid area,
and flatten it into the initial region token group:
\[
 U_{d,k}
 =
 \operatorname{Flatten}
 \left(\operatorname{Crop}(M_d,b_{d,k})\right).
\]
The number of tokens in $U_{d,k}$ depends on the spatial extent of the
candidate on the feature map.  All candidates reuse the same full-page feature
map and therefore require no additional vision-encoder pass.

\paragraph{Candidate-Private Contextualization.}
We append all region groups and 100 page-summary seeds after the base page
sequence.  The block-structured causal mask gives every region token access to
the base page and to preceding tokens in its own group, but blocks access to
sibling regions and summary tokens.  Summary tokens access the base page and
preceding summary tokens, but not region groups.  The base sequence cannot read
any appended state.  After the language model, a learned projection maps the
hidden states to 128 dimensions, followed by token-wise $\ell_2$
normalization.  Base-page hidden states are then discarded.  The projected
tokens of one group form its region entry, whereas the concatenation of all
valid region groups and the 100 summary-token outputs forms the compact page
entry.

\paragraph{Query Processing and Scoring.}
We use the same text processor as ColQwen2.5 and append ten query-augmentation
tokens after each query.  MaxSim is computed only
over valid, non-padding query and document vectors and is normalized by the
number of valid query tokens.
As defined in the main paper, the routing and deployed region score is
$F(q,d,k)=s_R(q,d,k)+s_D(q,d)$.  Chunking the candidates changes only memory
usage and does not alter this score.

\begin{table}[t]
    \centering
    \footnotesize
    \setlength{\tabcolsep}{3.5pt}
    \renewcommand{\arraystretch}{1.14}
    \begin{tabular}{@{}p{0.48\columnwidth}p{0.40\columnwidth}@{}}
        \toprule
        \textbf{Component} & \textbf{Setting} \\
        \midrule
        Retriever / auxiliary VLM initialization &
        Qwen2.5-VL-3B-Instruct \\
        Document embedding dimension & 128 \\
        Retriever LoRA & rank 64, $\alpha=128$ \\
        Auxiliary VLM LoRA & rank 32, $\alpha=64$ \\
        Page-summary tokens & 100 \\
        Region routing score & $s_R+s_D$ \\
        Generation target & Question followed by answer \\
        Training schedule & 5 epochs \\
        Learning rate & $1\times10^{-4}$ \\
        Per-GPU batch size & 24 \\
        Hardware & 16 NVIDIA A100-80GB GPUs \\
        NTP loss weight / routing temperature & $0.1 / 0.01$ \\
        Visual encoders & Frozen \\
        \bottomrule
    \end{tabular}
    \caption{Implementation settings for the full GARRET model. These
    hyperparameters are used throughout the experiments unless an ablation
    explicitly changes them.}
    \label{tab:supp_config}
\end{table}

\subsection{Generative-Feedback Implementation}

\subsubsection{Aligned Retrieval and Feedback Views}

Every candidate has two representations during training.  Its
\emph{retrieval view} is the page-contextualized token group produced by the
shared retriever, while its \emph{feedback view} is the exact pixel crop inside
the same canonical box.  To construct feedback crops, we convert the page to
RGB, resize it with Lanczos interpolation so that its shorter edge is 768
pixels, map the normalized box to the resized page, and use floor and ceiling
for the upper-left and lower-right boundaries, respectively.  The auxiliary
Qwen2.5-VL processor then resizes each crop subject to a minimum of
$4\times28\times28$ pixels and a maximum of $1024\times28\times28$ pixels.
This branch observes only pixels inside the selected box and therefore cannot
obtain evidence from a neighboring candidate.

\subsubsection{Local Candidate Pool}

The page-level retrieval loss uses distributed in-batch pages, but the
generative-feedback pool is constructed independently on each process.  For
query $q_i$, the pool contains its paired page, one hard-negative page, and one
random-negative page from the local minibatch.  We define the hardness of a
nonpaired page by the largest combined region score on that page,
\[
 h_i(d)=\max_k F(q_i,d,k), \qquad d\neq d_i^+,
\]
and select the page with the largest $h_i(d)$.  The random page is sampled from
the remaining nonpaired pages.  Once a page is selected, \emph{all} of its
valid layout candidates enter the pool, including distractors on the paired
page.  Therefore, the feedback mechanism learns both which page and which
within-page region support the target without treating any candidate as an
annotated positive.

\subsubsection{Stabilized Soft Routing}

Let $\mathcal C_i$ be the local candidate pool and let
$z_{i,r}=F(q_i,r)/\tau$ for $r\in\mathcal C_i$.  A very small routing
temperature can make the distribution nearly one-hot before the retriever has
learned meaningful region scores.  To preserve usable gradients, the
implementation caps the largest routing probability at $\rho=0.90$.  If the
ordinary softmax already satisfies the cap, we use a scale $\gamma_i=1$.
Otherwise, a detached binary search finds the largest
$\gamma_i\in(0,1)$ whose top probability does not exceed $\rho$.  The routing
weights are then
\[
 \pi_{i,r}
 =
 \frac{\exp(\gamma_i z_{i,r})}
 {\sum_{r'\in\mathcal C_i}\exp(\gamma_i z_{i,r'})}.
\]
The search for $\gamma_i$ is outside the gradient graph, but the final softmax
is evaluated on the original retrieval scores.  Gradients therefore propagate
through $F$ exactly as in standard differentiable routing.  This mechanism is
equivalent to increasing the effective temperature only for saturated rows;
it introduces no learned router parameters.

\subsubsection{Visual Source Key--Value Construction}

The auxiliary VLM encodes every candidate crop independently.  For each crop,
we retain only the key and value states at visual-token positions; instruction,
separator, and padding positions are excluded from the source memory.  Source
forwards are computed without gradient tracking, and the variable-length
visual memories are packed contiguously together with per-crop offsets.  This
avoids padding every crop to the largest visual-token count and ensures that
the routing weight for candidate $r$ controls only the visual states of that
candidate.

The source memories are constructed for every language layer of the auxiliary
VLM.  Correspondingly, retriever-weighted cross-attention is applied in
\emph{all auxiliary-VLM language layers}, rather than only in a subset of late
layers.  This all-layer design allows routed visual evidence to influence the
target representation throughout language processing.

\subsubsection{Retriever-Weighted All-Layer Cross-Attention}

The target branch contains the fixed instruction followed by the
teacher-forced generation target and contains no image tokens.  At every
language layer $\ell$, a target query state first attends to each crop source
separately:
\[
 c_{i,t,r}^{\ell}
 =
 \operatorname{Attn}
 \left(Q_{i,t}^{\ell},K_r^{\ell},V_r^{\ell}\right).
\]
Separate attention normalizations prevent visual tokens from different crops
from competing in a single token-level softmax.  To prevent candidates with
larger value-vector norms from receiving systematically larger contributions,
we divide each crop context by the attention-weighted value norm,
\[
\fitdisplay{\bar c_{i,t,r}^{\ell}
 =
 \frac{c_{i,t,r}^{\ell}}
 {\operatorname{Attn}
  \left(Q_{i,t}^{\ell},K_r^{\ell},\lVert V_r^{\ell}\rVert_2\right)
  +\epsilon},
 \qquad \epsilon=10^{-6}.}
\]
The retriever then combines the normalized crop contexts:
\[
\fitdisplay{e_{i,t}^{\ell}
 =
 \sum_{r\in\mathcal C_i}\pi_{i,r}\bar c_{i,t,r}^{\ell}.}
\]
The resulting context is added residually to the ordinary causal-attention
output before the layer output projection.  We compute this extra context only
at causal positions that predict supervised target tokens.  Because the source
K/V states are detached while $\pi_{i,r}$ remains differentiable, the path from
the language-model loss to the retriever is
\[
\fitdisplay{\mathcal L_{\mathrm{NTP}}
 \longrightarrow \pi_{i,r}
 \longrightarrow F(q_i,r)
 \longrightarrow \{s_R,s_D\}
 \longrightarrow \theta.}
\]
Candidate regions that reduce the prediction loss receive positive retrieval
feedback, whereas distractors receive the opposite update.  Complementary
regions may share routing mass and receive feedback jointly.

\subsubsection{Generation Target and Loss Reduction}

We use the fixed instruction
``Generate the question and its concise answer that are best supported by the
retrieved document regions,'' followed by the required output format.  The
supervised text is
\[
 y_i=
 \left[
 \texttt{Question:},\ q_i,\
 \texttt{Answer:},\ a_i
 \right].
\]
Instruction and assistant-prefix tokens are masked from the loss.  The
question--answer target is truncated to at most 100 supervised tokens; padding
tokens are also ignored.  If $\mathcal T_i$ denotes the remaining supervised
positions, the loss first averages token negative log-likelihoods within each
example and then averages over the local minibatch:
\[
\begin{aligned}
\mathcal L_{\mathrm{NTP}}
&=-\frac{1}{|\mathcal B|}\sum_{i\in\mathcal B}\frac{1}{|\mathcal T_i|}
\\[-2pt]
&\quad\times\sum_{t\in\mathcal T_i}\log p_{\phi,\theta}
\left(y_{i,t}\mid y_{i,<t},\mathcal I_i,\boldsymbol\pi_i\right).
\end{aligned}
\]
This per-example normalization prevents long questions or answers from
dominating the update.  GARRET uses the differentiable loss directly and never
converts auxiliary-VLM predictions into fixed positive or negative region
labels.

\subsubsection{Memory Bounds and Consistency Checks}

Crop sources are encoded in chunks of 32, and query--crop attention pairs are
evaluated in chunks of 64.  Before allocating source K/V states, the
implementation checks three bounds: at most 1,024 crop sources, at most 131,072
source visual tokens, and at most 8,388,608 products between supervised target
positions and routed source-token counts.  Exceeding a bound raises an error;
the implementation never silently drops candidates.  It also verifies that
region groups, pixel crops, source memories, candidate masks, and routing weights
have identical page and group indices.

\subsection{Optimization and Distributed Training}

The page objective is a ColPali-style batch-hard pairwise softplus loss.
Page representations are padded and gathered across
processes with gradients preserved, so each local query is contrasted against
pages from the distributed minibatch.  In contrast, the NTP candidate pool and
auxiliary-VLM computation remain local to each process.
Algorithm~\ref{alg:supp_train} summarizes one training step.  It emphasizes
the boundary between the distributed page
objective and the local feedback path: page representations are gathered for
retrieval supervision, whereas crop selection, visual-source construction, and
question--answer prediction use only candidates available on the current
process.  All candidates from the selected pages are retained, and no
query-to-region labels are introduced at any step.

\begin{algorithm}[t]
\caption{One GARRET training step}
\label{alg:supp_train}
\footnotesize
\begin{algorithmic}[1]
\REQUIRE Local batch $\mathcal B=\{(d_i^+,q_i,a_i)\}$; candidate sets
$\{\mathcal R(d)\}$
\STATE Encode each query and page; construct $T_R(d,k)$ and $T_D(d)$
\STATE Gather page representations and compute $\mathcal L_{\mathrm{page}}$
\STATE Initialize $\mathcal L_{\mathrm{NTP}}\leftarrow 0$
\FOR{each $(d_i^+,q_i,a_i)\in\mathcal B$}
    \STATE Select $d_i^+$, one hard-negative page, and one random-negative page
    \STATE Let $\mathcal C_i$ contain every valid candidate from these pages
    \STATE Compute $\{F(q_i,r)\}_{r\in\mathcal C_i}$ and stabilized
    $\{\pi_{i,r}\}_{r\in\mathcal C_i}$ with $\rho=0.90$
    \STATE Extract aligned pixel crops and detached layer-wise visual K/V
    \STATE Predict $y_i$ with retriever-weighted cross-attention in all layers
    \STATE Add the example-normalized token NLL to $\mathcal L_{\mathrm{NTP}}$
\ENDFOR
\STATE Average $\mathcal L_{\mathrm{NTP}}$ over $|\mathcal B|$
\STATE Optimize $\mathcal L=\mathcal L_{\mathrm{page}}
+0.1\mathcal L_{\mathrm{NTP}}$
\end{algorithmic}
\end{algorithm}

The detached source memories prevent gradients from flowing into crop-source
encoding, while the differentiable routing weights transmit the NTP signal to
the retriever.  Because the weighted cross-attention is inserted in every
auxiliary-VLM language layer, routed evidence can affect the complete target
forward pass.  The resulting training objective contains no additional
region-level supervision or diversity regularizer:
\[
 \mathcal L
 =
 \mathcal L_{\mathrm{page}}
 +0.1\,\mathcal L_{\mathrm{NTP}}.
\]

The retriever and auxiliary VLM are initialized from Qwen2.5-VL-3B-Instruct.
We freeze both vision encoders.  For the retriever, we train rank-64 LoRA
adapters with $\alpha=128$ in the language attention and MLP projections,
together with the retrieval projection.  For the auxiliary VLM, we train
rank-32 language-model LoRA adapters with $\alpha=64$; the auxiliary adapter
is synchronized across distributed workers.
Training uses BF16 model weights, SDPA attention, gradient checkpointing for
the retriever, 100 warm-up steps, a learning rate of $10^{-4}$, and a per-GPU
batch size of 24.  We train for five epochs on 16 A100-80GB GPUs.

\subsection{Deployment and Index Contents}

After training, we discard the auxiliary VLM, pixel-crop processor, source K/V
memories, training answers, candidate sampler, and NTP objective.  The deployed
system stores each contextualized region token group together with its page ID,
group ID, and bounding box.  The same region groups, concatenated with 100
page-summary embeddings, constitute the compact page entry.  Base-page hidden
states and padded positions are not indexed.  At inference, pages are ranked by
$s_D(q,d)$ and all persistent regions are ranked directly by
$F(q,d,k)=s_R(q,d,k)+s_D(q,d)$, without invoking the auxiliary VLM or applying
a hard page top-$K$ gate.

\subsubsection{Training Records Without Region Labels}

\begin{figure*}[p]
    \centering
    \includegraphics[width=\textwidth,height=0.88\textheight,
    keepaspectratio]
    {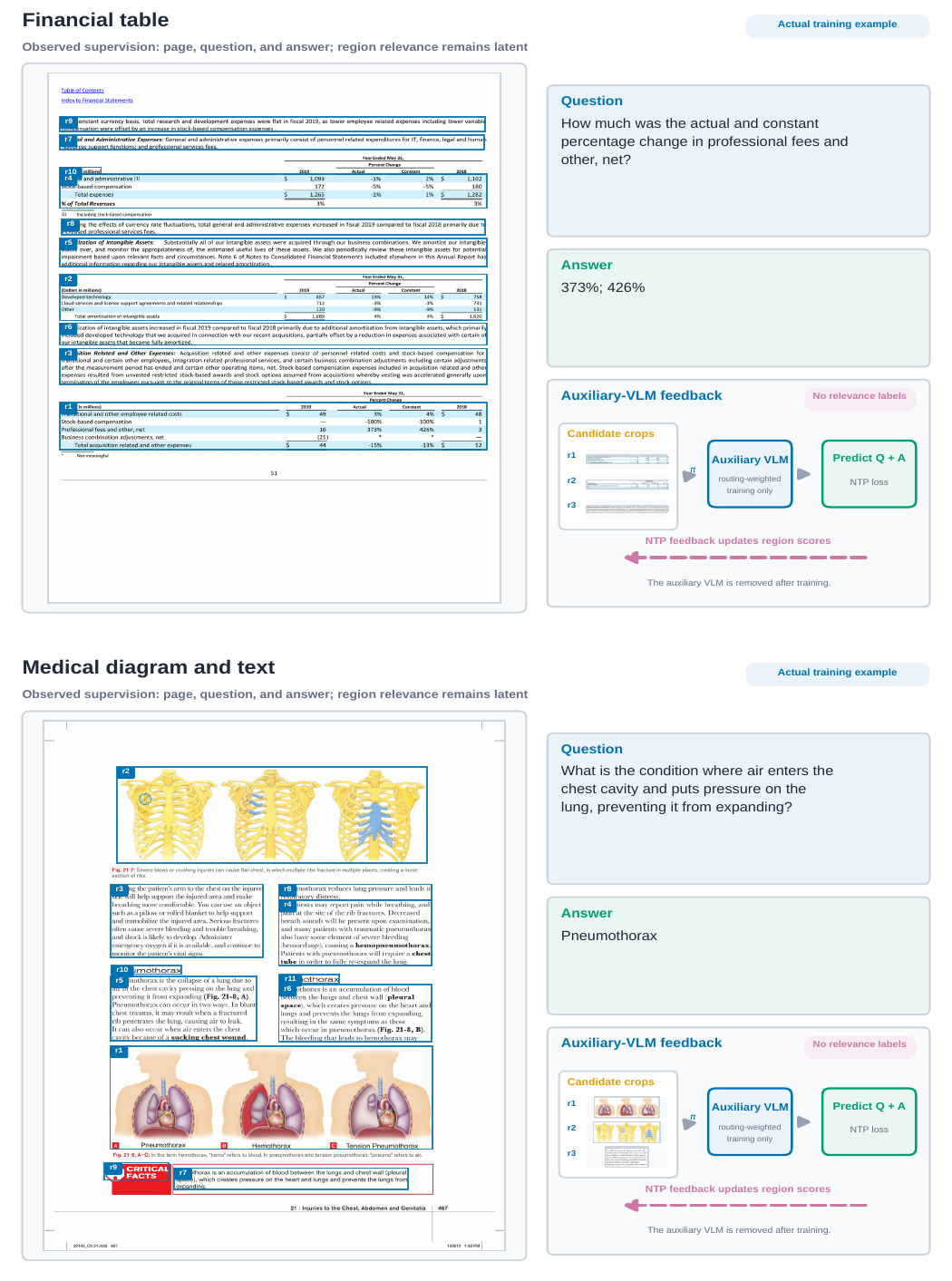}
    \caption{Two examples from the training set. Each training record contains
    only a page--query--answer triplet and query-independent layout bounding
    boxes generated by MinerU; no query-to-region relevance annotations are
    used.}
    \label{fig:supp_train_examples}
\end{figure*}

Figure~\ref{fig:supp_train_examples} shows two records from the training
collection. Each record pairs a page with a question and answer and retains
the page's predefined layout candidates. The candidate boxes define the
available evidence units, while the training tuple supplies no query-to-region
relevance label. The examples cover a financial table and a medical page with
diagrams and text. Both examples trace the training-only feedback path:
routing scores weight candidate crops for the auxiliary VLM, and the NTP loss
updates the region scores. GARRET removes the auxiliary VLM after training.

\section{ViDoRe-V3-Region Benchmark Construction}
\label{app:vidore-v3-region}

\subsection{Motivation and Source Data}

ViDoRe V3 evaluates retrieval at the page level and additionally provides
human-drawn evidence boxes on relevant pages.
These evidence boxes are free-form localization annotations: their boundaries
may cut through a semantic unit, merge several units, or contain substantial
surrounding context.  Directly treating them as retrievable regions would
therefore make the region corpus query-dependent and would not reflect a
realistic indexing setting.  We instead construct \textsc{ViDoRe-V3-Region}
by separating \emph{region proposal} from \emph{relevance annotation}.
Region proposals are extracted once from every page without access to any
query, while the original evidence boxes are used only to identify regions
that require relevance assessment.

We retain the complete query set and all eight public domains of ViDoRe V3:
human resources (HR), English financial documents (Fin-EN), industrial
documents, pharmaceutical documents, computer science (CS), energy, physics,
and French financial documents (Fin-FR).  Query text, page identifiers, and
the original query--page relevance associations are preserved.  The
construction procedure below is applied independently within each domain.

Table~\ref{tab:supp_data} summarizes the resulting evaluation collection.

\begin{table}[t]
    \centering
    \small
    \setlength{\tabcolsep}{6.0pt}
    \renewcommand{\arraystretch}{1.14}
    \begin{tabular}{lr}
        \toprule
        \textbf{Statistic} & \textbf{Value} \\
        \midrule
        Domains & 8 \\
        Pages & 19,252 \\
        Queries & 2,419 \\
        Fixed layout candidates & 200,682 \\
        \bottomrule
    \end{tabular}
    \caption{Statistics of \textsc{ViDoRe-V3-Region}. MinerU extracts 200,682
    query-independent layout candidates from 19,252 pages across eight
    domains.}
    \label{tab:supp_data}
\end{table}

\subsection{Query-Independent Layout-Region Corpus}

\paragraph{Layout Analysis.}
We run MinerU on every page image and collect the
bounding box of each detected layout element.  Importantly, MinerU receives
only the page and never the query or a ViDoRe evidence annotation.  Thus, for
a page \(p\), it produces a fixed set
\(\mathcal{B}_{p}=\{b_{p,1},\ldots,b_{p,n_p}\}\) that can be indexed before
any query is observed.

\paragraph{Box Validation and De-Duplication.}
We first remove boxes with non-positive width or height.  To suppress
redundant nested detections, the remaining boxes are sorted by area in
descending order; a box is discarded when at least \(90\%\) of its area is
covered by a larger box that has already been retained.  Duplicate boxes are
kept only once per page.  This filtering depends only on page layout and is
therefore independent of the benchmark queries.  Each surviving box is
assigned a stable region identifier.  If no valid layout box remains for a
page, a full-page box is used as a fallback so that the page remains
searchable.

After filtering, the union of the per-page sets forms a fixed corpus of
\(200{,}682\) layout regions.  This complete corpus, rather than only regions
on relevant pages, is indexed and ranked during evaluation.  Neither VLM
scoring nor human review changes the region boundaries or the set of region
identifiers.

\subsection{Mapping ViDoRe Evidence to Candidate Layout Regions}

For each official query--page relevance pair, we read the original ViDoRe V3
evidence boxes and express them in the same page coordinate system as the
MinerU regions.
Let \(\mathcal{A}_{q,p}\) denote the set of original evidence boxes for query
\(q\) on page \(p\).  A persistent MinerU region \(b\in\mathcal{B}_p\) is
selected for annotation if it has positive-area overlap with at least one
original evidence box:
\begin{equation}
    \mathcal{C}_{q,p}
    =
    \left\{
        b\in\mathcal{B}_{p}
        \ \middle|\
        \exists a\in\mathcal{A}_{q,p},
        \operatorname{IoU}(a,b)>0
    \right\},
    \label{eq:candidate-regions}
\end{equation}
where
\begin{equation}
    \operatorname{IoU}(a,b)
    =
    \frac{\operatorname{area}(a\cap b)}
         {\operatorname{area}(a)+\operatorname{area}(b)
          -\operatorname{area}(a\cap b)}.
\end{equation}
If a layout region overlaps multiple evidence boxes, it is included only
once for that query--page pair.  We deliberately use the permissive
\(\operatorname{IoU}>0\) rule at this stage to favor recall: an imprecise
free-form evidence boundary should not prevent a semantically correct layout
unit from reaching the subsequent semantic and human checks.  Conversely,
geometric overlap alone never determines the final label.

\begin{table*}[t]
    \centering
    \begingroup
    \setlength{\fboxsep}{6pt}

    \textbf{Joint Question--Answer Target (GARRET)}\\[3pt]
    \fbox{%
    \begin{minipage}{0.94\textwidth}
        \footnotesize\ttfamily\raggedright
        Generate the question and its concise answer that are best supported
        by the retrieved document regions. Use exactly this format:\\
        Question: <question>\\
        Answer: <answer>
    \end{minipage}}\\[7pt]

    \textbf{Question Target Only}\\[3pt]
    \fbox{%
    \begin{minipage}{0.94\textwidth}
        \footnotesize\ttfamily\raggedright
        Generate the search query that is best supported by the retrieved
        document regions.
    \end{minipage}}\\[7pt]

    \textbf{Answer Target Only}\\[3pt]
    \fbox{%
    \begin{minipage}{0.94\textwidth}
        \footnotesize\ttfamily\raggedright
        Complete the concise answer after the provided question prefix using
        the retrieved document regions. Continue after the Answer: prefix with
        the answer only.
    \end{minipage}}

    \caption{Exact training prompts for the generative-target ablations. The
    variants change only the supervised generation target while retaining the
    same retriever, candidate pool, and page-level objective.}
    \label{tab:supp_target_prompts}
    \endgroup
\end{table*}

\begin{table}[t]
    \centering
    \footnotesize
    \setlength{\tabcolsep}{5.0pt}
    \renewcommand{\arraystretch}{1.14}
    \begin{tabular}{@{}lccc@{}}
        \toprule
        \textbf{Summary tokens} & \textbf{Vectors/page} &
        \textbf{Region} & \textbf{Page} \\
        \midrule
        0   & 292.9 & 40.06 & 55.82 \\
        50  & 342.9 & 40.95 & 56.59 \\
        \textbf{100} & 392.9 & \underline{41.26} & \textbf{57.29} \\
        150 & 442.9 & \textbf{41.30} & \underline{57.20} \\
        \bottomrule
    \end{tabular}
    \caption{Effect of the page-summary-token budget on ViDoRe V3 nDCG@10
    (\(\times100\)). We use 100 tokens in the full model, as indicated by the
    bold token budget. Bold and underlined metric values denote the best and
    second-best results, respectively.}
    \label{tab:supp_summary_length}
\end{table}

\begin{table}[t]
    \centering
    \footnotesize
    \setlength{\tabcolsep}{4.0pt}
    \renewcommand{\arraystretch}{1.14}
    \begin{tabular}{@{}cccrr@{}}
        \toprule
        \textbf{Hard} & \textbf{Random} & \textbf{Total pages} &
        \textbf{Region} & \textbf{Page} \\
        \midrule
        0 & 1 & 2 & 38.51 & 57.08 \\
        0 & 2 & 3 & 39.42 & \textbf{57.32} \\
        1 & 0 & 2 & \underline{40.39} & 57.15 \\
        2 & 0 & 3 & 40.01 & 57.21 \\
        \textbf{1} & \textbf{1} & \textbf{3} & \textbf{41.26} &
        \underline{57.29} \\
        \bottomrule
    \end{tabular}
    \caption{Ablation of negative-page sampling for the local candidate pool
    on ViDoRe V3 nDCG@10 (\(\times100\)). ``Total pages'' includes the paired
    page, which is present in every row. The full model uses one hard-negative
    and one random-negative page.}
    \label{tab:supp_candidate_sampling}
\end{table}

\subsection{VLM-Assisted Relevance Pre-Annotation}

\paragraph{Visual Input.}
Each candidate in \(\mathcal{C}_{q,p}\) is independently assessed in a
separate VLM inference call.  We draw \emph{exactly one} candidate box in red on the full page
for each model input.  The model therefore scores only one region at a time;
an input never contains multiple candidate boxes.  This design requires many
inference calls, but reduces ambiguity about which visual region is being
judged.
The full page is retained to preserve layout context, while the prompt
explicitly requires the score to be based only on content inside the single
red box.

\paragraph{Model and Inference.}
We use Qwen3-VL-32B-Instruct.  The model receives the
single-box page image and the corresponding query.  Before producing the
score, it is required to generate a short reasoning statement that explicitly
checks whether the evidence is contained inside the highlighted region.  This
rationale-first procedure is intended to encourage deliberate region--query
comparison.  The exact prompt is reproduced in
Table~\ref{tab:supp_region_relevance_prompt}.

\subsection{Human Verification}

The VLM score and its accompanying reasoning are provided to two annotators
as references.  The annotators independently review every candidate region
and assign a binary relevance label.  A region is considered relevant only
when the content inside the box provides direct or indispensable evidence for
answering the query.  Regions with only topical similarity, incidental
geometric overlap, or evidence located primarily outside the box are
rejected.  Disagreements are discussed until consensus is reached, and the
consensus binary labels form the final qrels. These qrels are evaluation-only
annotations and are never used to train GARRET.

\begin{figure*}[p]
    \centering
    \includegraphics[width=\textwidth,height=0.80\textheight,
    keepaspectratio]
    {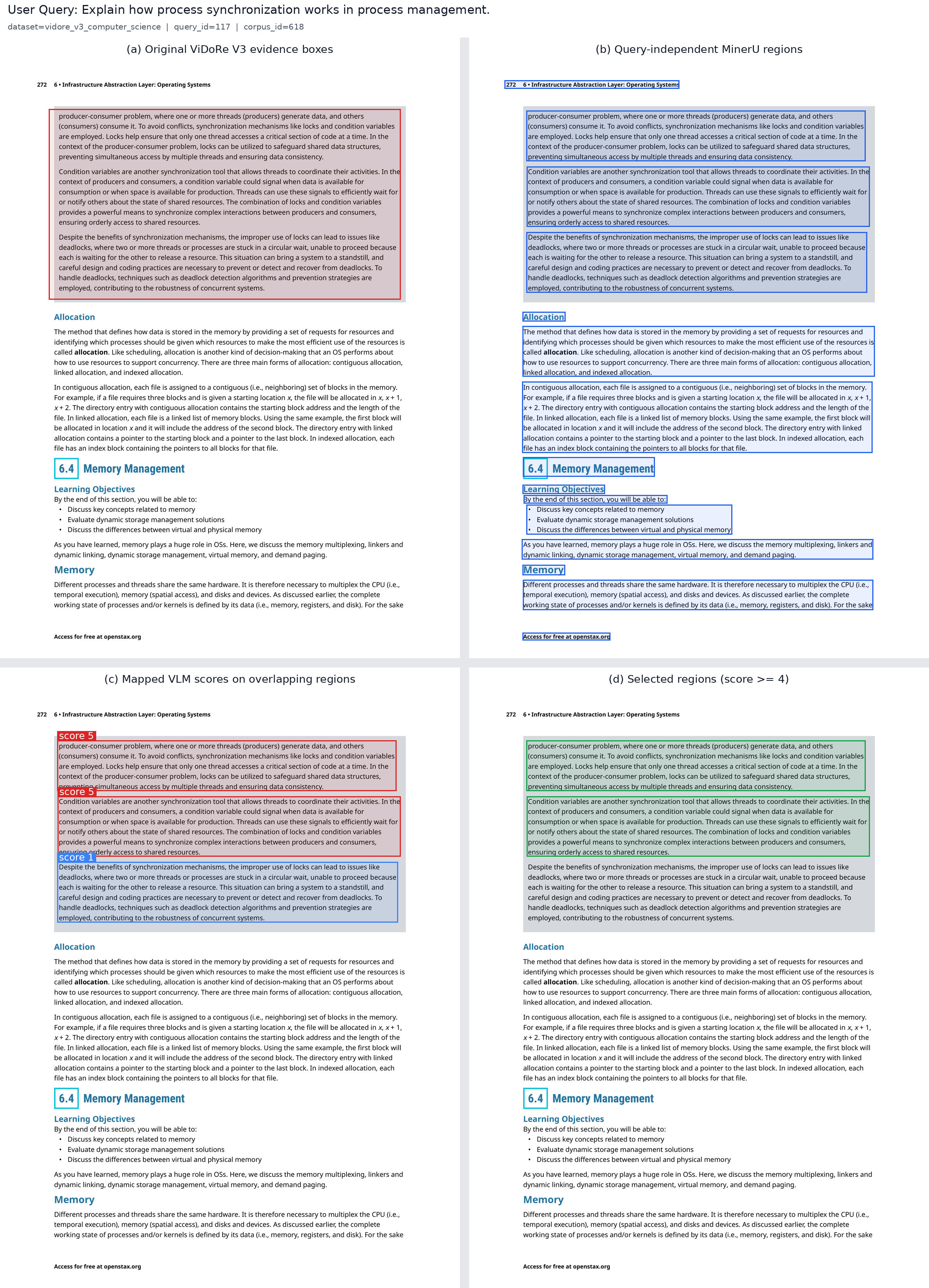}
    \caption{ViDoRe-V3-Region construction for an example query.
    (a) The original ViDoRe V3 annotation is a free-form evidence box covering
    three paragraphs. (b) MinerU, run without access to the query, divides the
    page into persistent layout regions. (c) The VLM independently scores each
    region that intersects the original evidence box. (d) Green boxes mark
    regions with scores of at least 4 as provisional positives. Two human
    annotators subsequently review every scored candidate and determine the
    final qrels.}
    \label{fig:supp_benchmark_construction}
\end{figure*}

Figure~\ref{fig:supp_benchmark_construction} visualizes the complete process,
from the original free-form evidence box to
query-independent proposals, per-region VLM scores, and the provisional
regions passed to human verification.

\section{Additional Ablation Studies}
\label{sec:supp_additional_ablations}

We provide the exact training prompts for the generative-target controls and
further ablate two design choices held fixed in the main experiments: the
number of page-summary tokens and the composition of the local candidate
pool. Unless explicitly varied, all models use the training data,
architecture, initialization, optimization schedule, and ViDoRe V3 evaluation
protocol of the full GARRET model. Region and page results are macro
nDCG@10 scores ($\times 100$).

\subsection{Prompts for the Generative-Target Controls}

Table~\ref{tab:supp_target_prompts} gives the instructions for the three target
variants verbatim. The fixed instruction for joint question--answer and
question-only generation contains neither the ground-truth question nor its
answer. For answer-only generation, the question is supplied as a
teacher-forced prefix so that the loss measures whether the routed evidence
supports the answer to that particular question.

For the joint target, teacher forcing serializes the target as
\(\texttt{Question:}\ q_i\ \texttt{Answer:}\ a_i\), and the NTP loss covers
both question and answer tokens. The question-only control predicts only
\(q_i\). The answer-only control receives
\(\texttt{Question:}\ q_i\ \texttt{Answer:}\) as the prefix and applies the
loss only to the continuation \(a_i\). Thus, the controls preserve the same
retrieval architecture, candidate pool, and page-level objective while
isolating the supervision supplied by the question and answer tokens.

\begin{table*}[t]
    \centering
    \footnotesize
    \setlength{\tabcolsep}{4.0pt}
    \renewcommand{\arraystretch}{1.08}
    \begin{tabular}{@{}p{0.16\textwidth}p{0.19\textwidth}
                       p{0.34\textwidth}p{0.17\textwidth}@{}}
        \toprule
        \textbf{Method family} & \textbf{Candidate input} &
        \textbf{Candidate representation} & \textbf{Region score} \\
        \midrule
        BM25 &
        Parsed candidate text &
        Sparse lexical index &
        Lexical matching score \\
        BGE-M3 / Qwen3-Embedding &
        Parsed candidate text &
        One dense text vector &
        Dense similarity \\
        ColPali / ColQwen &
        Exact candidate crop &
        Independently encoded visual vectors &
        Late interaction \\
        RegionRAG &
        Query-dependent predicted regions &
        Dynamic region output mapped to benchmark boxes &
        Mapped region score \\
        GARRET &
        Full page and candidate box &
        Contextual region token group &
        $s_R+s_D$ \\
        \bottomrule
    \end{tabular}
    \caption{Adaptation of each retriever to fixed-candidate region ranking.
    Text retrievers encode parsed candidate text, ColPali and ColQwen encode
    isolated candidate crops, and RegionRAG maps its query-dependent outputs
    to benchmark boxes. GARRET directly scores persistent, page-contextualized
    candidates.}
    \label{tab:supp_baselines}
\end{table*}

\subsection{Number of Page-Summary Tokens}

Page-summary tokens provide a compact global readout in addition to the
contextualized region groups. Table~\ref{tab:supp_summary_length} varies their
number while leaving the attention pattern and all other settings unchanged.
Because each summary token contributes one document vector, the setting with
\(n_s\) summary tokens indexes \(292.9+n_s\) vectors per page on average.

Increasing the budget from 0 to 100 improves region and page nDCG@10 by 1.20
and 1.47 points, respectively, confirming that the summary readout captures
useful page information not fully covered by the fixed layout candidates.
The gains saturate after 100 tokens: using 150 tokens changes region nDCG@10
by only +0.04 and page nDCG@10 by -0.09, while adding 50 vectors per page.
We therefore select 100 tokens as the better quality--index-size trade-off.

\subsection{Candidate-Pool Composition}

The retriever-weighted cross-attention branch processes every valid candidate
from each selected page. Its training memory therefore grows with the number
of pages in the local pool. Under our hardware configuration, the pool can
contain at most three pages per example. The paired page is always included;
we vary how the remaining one or two slots are allocated between hard-negative
pages and randomly sampled negative pages. A hard-negative page is selected by
its highest-scoring candidate, as defined in the main paper, whereas a random
page is sampled from the nonpaired pages outside the hard-negative set.

With one negative page, hard-negative sampling is substantially more useful
for region learning than random sampling (40.39 versus 38.51), because it
exposes the router to more confusable candidates. Filling both available
negative slots exclusively with hard negatives does not improve further,
while the balanced allocation achieves the best region result, 41.26. This
suggests complementary roles: hard negatives sharpen fine-grained
discrimination and random negatives broaden negative coverage. Across all
allocations, page nDCG@10 remains within a narrow 0.24-point range, whereas
region nDCG@10 spans 2.75 points. Candidate-pool composition therefore mainly
affects the fine-grained generative-feedback signal rather than the separately
supervised page objective.

\section{Evaluation Protocols}

\subsection{ViDoRe-V3-Region Benchmark}

The fixed candidate corpus and human-verified binary qrels are constructed as
described in \emph{ViDoRe-V3-Region Benchmark Construction} section. At evaluation time, each
method ranks those same region identifiers corpus-wide for every query; no
construction-time evidence box, VLM score, or annotation is exposed to the
retriever.

The benchmark ranks all fixed candidates across the corpus. For a query $q$,
binary relevance $\operatorname{rel}_q(j)$ at rank $j$ gives
\[
\operatorname{DCG@10}(q)=
\sum_{j=1}^{10}
\frac{\operatorname{rel}_q(j)}{\log_2(j+1)}.
\]
We divide by the ideal DCG for the query to obtain nDCG@10. We first average
queries within each domain and then report the unweighted mean across the
eight domains. Page-level and region-level nDCG use different retrieval units
and qrels; their absolute values describe different tasks.

\subsection{MMDocIR Protocol}

MMDocIR restricts each query to candidates from its
known source document. The resulting task measures within-document evidence
localization rather than corpus-wide retrieval.

\paragraph{Official Layout Recall.}
We first report the benchmark's overlap-based Layout Recall. For each query,
the evaluator sums the intersection area between every top-$K$ candidate and
every gold box on the same page, and divides this sum by the total area of all
gold boxes for that query. Because intersections are accumulated independently,
overlapping retrieved candidates may contribute repeatedly. We report Layout
R@1, R@5, and R@10, averaged per query over all 1,658 queries.

\paragraph{Project-Derived Binary nDCG.}
We additionally construct frozen binary relevance judgments over the official
layout candidates. For a box $b=(x_1,y_1,x_2,y_2)$ recorded on a page of width
$W$ and height $H$, we convert it to page-normalized coordinates as
\[
\mathcal{N}(b;W,H)
=
\left(x_1/W,\;y_1/H,\;x_2/W,\;y_2/H\right).
\]
For candidate $r$ and gold box $g$, let
$\bar b_r=\mathcal{N}(b_r;W_r,H_r)$ and
$\bar b_g=\mathcal{N}(b_g;W_g,H_g)$. We compute standard intersection over
union, using the union area as the denominator, when the two boxes belong to
the same page:
\[
\fitdisplay{I(r,g)
=
\begin{cases}
\frac{
\operatorname{area}\!\left(\bar b_r\cap\bar b_g\right)
}{
\operatorname{area}(\bar b_r)+\operatorname{area}(\bar b_g)
-\operatorname{area}(\bar b_r\cap\bar b_g)
}, & p_r=p_g,\\[6pt]
0, & p_r\neq p_g.
\end{cases}}
\]
For the gold-box set $\mathcal{G}_q$ of query $q$, binary relevance is
\[
\fitdisplay{\operatorname{rel}_q(r)
=
\mathbf{1}\!\left[
\max_{g\in\mathcal{G}_q} I(r,g)\geq 0.5
\right].}
\]
This is an any-match rather than one-to-one mapping:
multiple candidates may match the same gold box, and a candidate may overlap
multiple gold boxes.

The mapping is computed once and reused unchanged for every method. Across
170,338 candidates and 2,638 gold boxes, it yields 2,526 positive
query--region pairs. We compute nDCG@5 and nDCG@10 on the 1,600 of 1,658
queries having at least one mapped positive; the remaining 58 queries have
zero ideal DCG under this derived mapping and are excluded. Rankings are
ordered by decreasing retrieval score, with exact ties broken
lexicographically by the stable region identifier. We first average per-query
scores within each of the ten domains and then report their unweighted macro
average. These nDCG measures are project-defined, rank-sensitive extensions
and are not official MMDocIR metrics.

\subsection{Baseline Adaptations and Index Accounting}

\subsubsection{Fixed-Candidate Baselines}

Table~\ref{tab:supp_baselines} specifies how each model produces a score for a
fixed candidate. Every ViDoRe-V3-Region method ranks the same candidate
identifiers. The text baselines use BM25, BGE-M3, and Qwen3-Embedding. The
structural controls follow the corresponding LFRAG and RegionRAG designs.

\subsubsection{Indexed-Vector Counts}

We count non-padding document-side vectors after the retrieval projection.
For an independently encoded crop, the region count equals the number of
visual retrieval vectors for that crop. For GARRET, the region count equals
$|Z_{d,k}|$, and the compact page count equals
\[
\sum_{k=1}^{m_d}|Z_{d,k}|+|S_d|.
\]
We average region counts over candidates and page counts over pages. Text
single-vector models use one dense vector per item; BM25 has no comparable
dense-vector count.

The count measures representation length. Separate measurements capture
embedding bytes, search latency, FLOPs, and peak memory. GARRET reuses the same
computed region states when it constructs region and page views. The reported
representation length counts each reused GARRET region vector once and does
not imply a particular physical index implementation.

\subsection{Downstream RAG Protocol}

\subsubsection{Generation and Judging}

The downstream experiment uses the ViDoRe V3 test set. All methods use the
same Qwen2.5-VL-7B-Instruct generator, decoding configuration, answer prompt,
and image preprocessing. We use deterministic
decoding with a maximum of 512 new tokens. We convert each image to RGB and
cap its longest side at 768 pixels.

Page-5 supplies the five highest-ranked full-page images. GARRET supplies the
top-$k$ candidate crops for $k\in\{5,10,15,20\}$. The generator receives the
crops as separate visual inputs; training answers remain excluded. We assign
evidence identifiers in retrieval order and place the corresponding images
before the text prompt.

A text-only Qwen2.5-VL-32B-Instruct judge receives the question, canonical
reference answer, and parsed model answer. The judge assigns
\texttt{Correct}, \texttt{Partially Correct}, or \texttt{Incorrect}. Strict
accuracy counts only \texttt{Correct}; all other labels and invalid outputs
receive zero, and all queries remain in the denominator.

\subsubsection{RAG Prompt Templates}

Table~\ref{tab:supp_rag_prompts} reports the complete evidence-to-answer and
correctness-judge templates. Braced fields denote values instantiated for
each query.

\subsubsection{Retrieved-Area Accounting}

For query $q$, let $\mathcal P_5(q)$ contain the five retrieved pages and let
$\mathcal R_k(q)$ contain the $k$ retrieved regions. We define
\[
\operatorname{Area/P5}(q)=
100\,
\frac{\sum_{r\in\mathcal R_k(q)}\operatorname{area}(r)}
{\sum_{p\in\mathcal P_5(q)}\operatorname{area}(p)}.
\]
The table in the main paper reports the mean of this query-level ratio.
Because the generator receives every crop separately, the numerator counts
overlapping crops separately. Area/P5 measures retrieved image area. It does
not measure generator tokens, latency, or FLOPs.

\subsection{Evaluation Scope}

GARRET ranks regions from a fixed candidate parser. A missed layout candidate
therefore limits downstream region recall even when the retriever scores every
available candidate correctly. The ViDoRe-V3-Region results measure ranking
within this fixed candidate collection.

The region qrels use Qwen3-VL scores to initialize candidate labels and two
annotators to verify all labels. We keep the resulting binary judgments
separate from the page qrels and from the MMDocIR localization metrics.

The auxiliary VLM increases training cost. GARRET removes it after training,
so deployed page and region retrieval use only the query encoder and
precomputed document-side vectors. The vector counts reported in the paper
describe representation length; end-to-end indexing time, byte-level storage,
and query latency require separate measurements.

\section{More Results}

This section provides additional quantitative results and qualitative
retrieval analyses.

\subsection{Quantitative Results}

\subsubsection{Per-Domain Results with Generative Feedback}

Table~\ref{tab:supp_feedback_domains} expands the macro ablation from the main
paper. Removing the auxiliary VLM and NTP path lowers region nDCG@10 by 4.93
points and page nDCG@10 by 0.25 points. The larger region change shows that
generative feedback primarily improves fine-grained candidate ranking.

\begin{table*}[tp]
{
\centering
    \footnotesize
    \setlength{\tabcolsep}{2.5pt}
    \renewcommand{\arraystretch}{1.06}
    \begin{tabular*}{\textwidth}{
        @{\extracolsep{\fill}}lp{0.17\textwidth}*{9}{r}@{}
    }
        \toprule
        \textbf{View} & \textbf{Variant} & \textbf{Avg.} & \textbf{CS} &
        \textbf{Ene.} & \textbf{Fin-EN} & \textbf{Fin-FR} & \textbf{HR} &
        \textbf{Ind.} & \textbf{Pha.} & \textbf{Phy.} \\
        \midrule
        Region & GARRET
        & 41.26 & 55.46 & 42.47 & 34.05 & 22.37 & 46.07 & 41.93 & 49.72 & 38.04 \\
        Region & No generative feedback
        & 36.33 & 50.96 & 34.92 & 34.43 & 19.38 & 39.78 & 34.63 & 44.73 & 31.80 \\
        \addlinespace[2pt]
        Page & GARRET
        & 57.29 & 76.46 & 60.44 & 59.10 & 39.26 & 61.06 & 52.71 & 63.04 & 46.24 \\
        Page & No generative feedback
        & 57.04 & 76.27 & 59.66 & 62.21 & 37.58 & 61.02 & 49.48 & 64.57 & 45.51 \\
        \bottomrule
    \end{tabular*}
    \caption{Per-domain ablation of generative feedback on ViDoRe V3 using
    nDCG@10 ($\times 100$). Removing the auxiliary VLM and NTP objective
    reduces average region retrieval by 4.93 points but page retrieval by only
    0.25 points, showing that generative feedback primarily improves
    fine-grained ranking. Values are point estimates from one trained model
    per variant.}
    \label{tab:supp_feedback_domains}
}
\end{table*}

\subsubsection{Per-Subset Page Retrieval}

Tables~\ref{tab:supp_page_v1}--\ref{tab:supp_page_domains} expand the page
averages from the main paper. All tables report nDCG@10, and the average is
the unweighted mean across subsets. ViDoRe V1 uses all published queries,
ViDoRe V2 uses all published languages, and ViDoRe V3 uses its eight page
subsets. Page and region judgments define different retrieval tasks.

\begin{table*}[tp]
{
\centering
    \footnotesize
    \setlength{\tabcolsep}{1.35pt}
    \renewcommand{\arraystretch}{1.06}
    \begin{tabular*}{\textwidth}{
        @{\extracolsep{\fill}}l*{11}{r}@{}
    }
        \toprule
        \textbf{Method} & \textbf{Avg.} & \textbf{ArXiv} & \textbf{DocV} &
        \textbf{InfoV} & \textbf{Shift} & \textbf{Syn-AI} & \textbf{Syn-En} &
        \textbf{Syn-Gov} & \textbf{Syn-HC} & \textbf{TabFQ} & \textbf{TatD} \\
        \midrule
        \multicolumn{12}{l}{\textit{Text-based document retrievers}} \\
        BM25
        & 69.94 & 36.65 & 39.45 & 68.36 & 68.29 & 95.85 & 90.90 & 86.52
        & 90.93 & 61.67 & 60.73 \\
        BGE-M3
        & 68.27 & 32.83 & 29.50 & 67.95 & 73.10 & 90.56 & 82.71 & 84.81
        & 91.16 & 89.69 & 40.38 \\
        Qwen3-Emb.-8B
        & 75.82 & 46.05 & 36.58 & 81.81 & 77.73 & 95.44 & 90.02 & 90.84
        & 93.93 & \textbf{93.92} & 51.89 \\
        \midrule
        \multicolumn{12}{l}{\textit{Vision-based document retrievers}} \\
        ColPali-v1.3
        & 85.42 & 84.36 & 59.86 & 86.83 & 77.51 & 97.27 & 94.87 & 95.79
        & 96.47 & 88.44 & 72.77 \\
        ColQwen2.5-v0.2
        & \underline{88.45} & \underline{88.67} & \textbf{61.97}
        & \underline{91.51} & \underline{85.89} & \underline{99.13}
        & 95.45 & 95.23 & 97.29 & 89.09 & \underline{80.25} \\
        RegionRAG
        & 86.70 & 81.52 & 56.05 & 89.89 & \textbf{88.53}
        & \underline{99.13} & \underline{96.63} & \underline{96.02}
        & \underline{98.89} & \underline{92.57} & 67.79 \\
        \textbf{GARRET}
        & \textbf{88.56} & \textbf{90.62} & \underline{60.34}
        & \textbf{91.89} & 85.84 & \textbf{100.00} & \textbf{96.65}
        & \textbf{97.66} & \textbf{99.26} & 82.34 & \textbf{80.99} \\
        \bottomrule
    \end{tabular*}
    \caption{Per-subset page retrieval on ViDoRe V1 using nDCG@10
    ($\times 100$). The Avg. column is the unweighted mean across the ten
    published subsets. Bold and underlined values denote the best and
    second-best results.}
    \label{tab:supp_page_v1}
}
\end{table*}

\begin{table*}[tp]
{
\centering
    \footnotesize
    \setlength{\tabcolsep}{5.0pt}
    \renewcommand{\arraystretch}{1.06}
    \begin{tabular}{@{}lrrrrr@{}}
        \toprule
        \textbf{Method} & \textbf{Avg.} & \textbf{Biomedical} &
        \textbf{Economics} & \textbf{ESG-Human} & \textbf{ESG} \\
        \midrule
        \multicolumn{6}{l}{\textit{Text-based document retrievers}} \\
        BM25 & 21.61 & 15.79 & 16.28 & 35.58 & 18.79 \\
        BGE-M3 & 49.75 & 44.71 & 48.79 & 58.27 & 47.21 \\
        Qwen3-Emb.-8B & 55.28 & 59.56 & 50.98 & 59.61 & 50.96 \\
        \midrule
        \multicolumn{6}{l}{\textit{Vision-based document retrievers}} \\
        ColPali-v1.3 & 57.81 & 59.91 & 49.42 & \underline{63.04}
        & \textbf{58.85} \\
        ColQwen2.5-v0.2 & \underline{58.59} & \underline{60.64} & 52.90
        & \textbf{63.70} & \underline{57.12} \\
        RegionRAG & \textbf{59.85} & \textbf{62.51} & \textbf{58.20}
        & 62.27 & 56.41 \\
        \textbf{GARRET} & 57.51 & 60.27 & \underline{52.99} & 62.79 & 53.98 \\
        \bottomrule
    \end{tabular}
    \caption{Per-subset page retrieval on ViDoRe V2 using nDCG@10
    ($\times 100$) under the published all-language protocol. The Avg. column
    is the unweighted mean across the four subsets. Bold and underlined values
    denote the best and second-best results.}
    \label{tab:supp_page_v2}
}
\end{table*}

\begin{table*}[tp]
{
\centering
    \footnotesize
    \setlength{\tabcolsep}{2.5pt}
    \renewcommand{\arraystretch}{1.06}
    \begin{tabular*}{\textwidth}{
        @{\extracolsep{\fill}}l*{9}{r}@{}
    }
        \toprule
        \textbf{Method} & \textbf{Avg.} & \textbf{CS} & \textbf{Ene.} &
        \textbf{Fin-EN} & \textbf{Fin-FR} & \textbf{HR} & \textbf{Ind.} &
        \textbf{Pha.} & \textbf{Phy.} \\
        \midrule
        \multicolumn{10}{l}{\textit{Text-based document retrievers}} \\
        BM25
        & 38.29 & 63.89 & 15.64 & 50.75 & 11.87 & 49.35 & 45.76 & 55.27 & 13.75 \\
        BGE-M3
        & 44.63 & 63.63 & 50.09 & 43.86 & 24.24 & 45.35 & 39.02 & 54.63 & 36.21 \\
        Qwen3-Emb.-8B
        & 53.93 & 73.78 & 59.09 & 56.49 & 36.01 & 53.70 & 44.94 & 62.11 & 45.31 \\
        \midrule
        \multicolumn{10}{l}{\textit{Vision-based document retrievers}} \\
        ColPali-v1.3
        & 48.86 & 73.14 & 46.09 & 45.02 & 22.61 & 54.57 & 47.89 & 59.38 & 42.16 \\
        ColQwen2.5-v0.2
        & \underline{57.18} & 76.07 & \underline{60.12} & \underline{61.85}
        & 37.85 & \underline{59.54} & \underline{52.79} & \textbf{63.08}
        & \underline{46.14} \\
        RegionRAG
        & 57.14 & \textbf{76.82} & 58.78 & \textbf{61.93} & \textbf{39.89}
        & 58.20 & \textbf{53.67} & 62.69 & 45.17 \\
        \textbf{GARRET}
        & \textbf{57.29} & \underline{76.46} & \textbf{60.44} & 59.10
        & \underline{39.26} & \textbf{61.06} & 52.71 & \underline{63.04}
        & \textbf{46.24} \\
        \bottomrule
    \end{tabular*}
    \caption{Per-domain page retrieval on ViDoRe V3 using nDCG@10
    ($\times 100$) and the official page judgments. The Avg. column is the
    unweighted mean across the eight domains. Bold and underlined values
    denote the best and second-best results.}
    \label{tab:supp_page_domains}
}
\end{table*}

\subsection{Qualitative Results}

\subsubsection{Ranked Evidence Units}

\begin{figure*}[p]
    \centering
    \includegraphics[width=\textwidth,height=0.80\textheight,
    keepaspectratio]
    {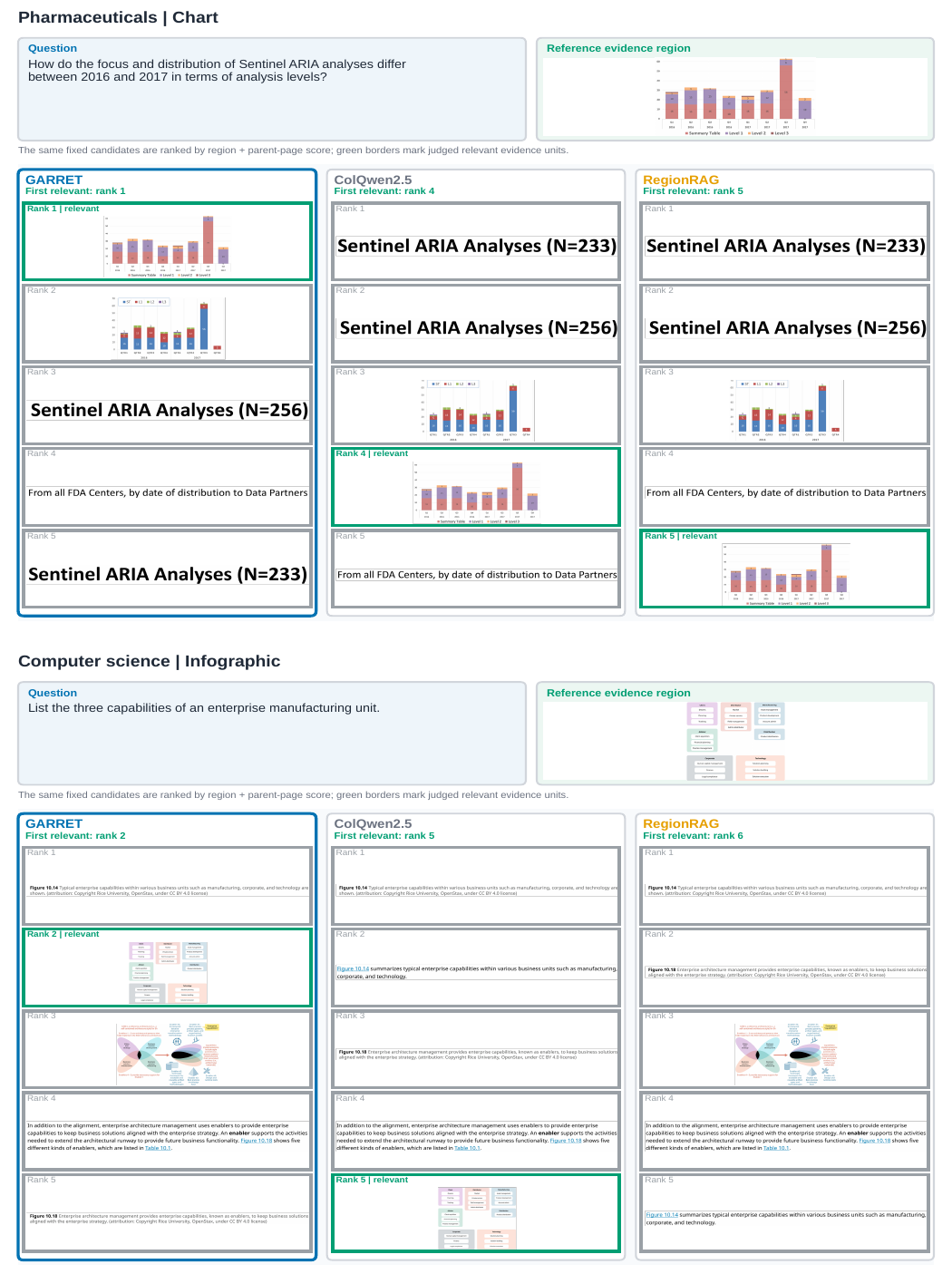}
    \caption{Top-ranked fixed candidates from GARRET, ColQwen2.5, and
    RegionRAG. \emph{Top}: for an example from the pharmaceuticals subset of
    ViDoRe V3, the first relevant candidate appears at rank 1 for GARRET, rank
    4 for ColQwen2.5, and rank 5 for RegionRAG. \emph{Bottom}: for an example
    from the computer-science subset, the first relevant candidate appears at
    rank 2 for GARRET, rank 5 for ColQwen2.5, and rank 6 for RegionRAG.}
    \label{fig:supp_retrieval_examples}
\end{figure*}

Figure~\ref{fig:supp_retrieval_examples} compares GARRET, ColQwen2.5, and
RegionRAG on the same fixed candidate corpus. Each method combines its region
score with the score of the candidate's parent page. The upper-right panel in
each example shows the reference evidence region, and green frames mark
retrieved candidates judged relevant under the evaluation qrels.

For the pharmaceutical query, GARRET places the reference chart at rank 1,
compared with ranks 4 and 5 for ColQwen2.5 and RegionRAG. For the
computer-science query, GARRET places the reference infographic at rank 2,
compared with ranks 5 and 6.

\subsubsection{Full-Page Spatial Similarity}
\begin{figure*}[p]
    \centering
    \includegraphics[width=\textwidth]
    {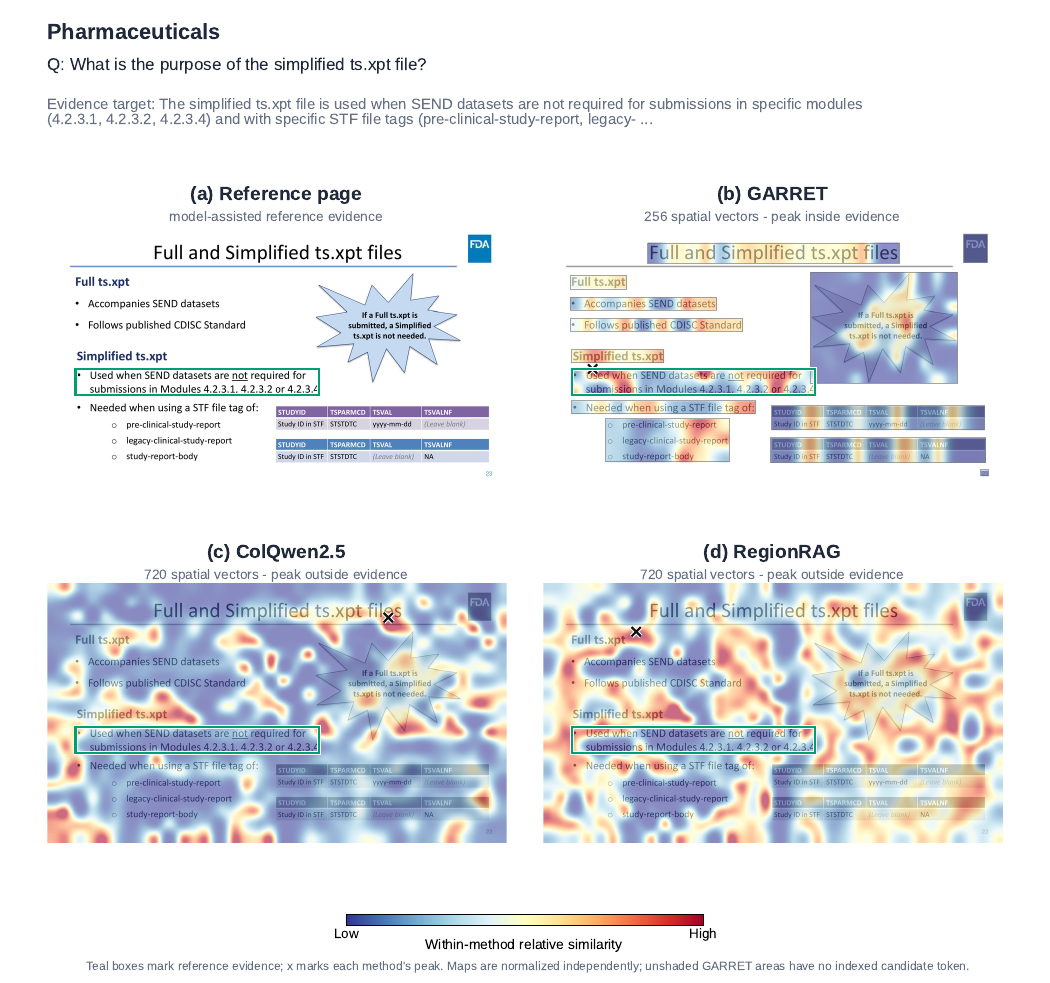}
    \caption{Query--page spatial similarity for an example from the
    pharmaceuticals subset of ViDoRe V3. Panel (a) shows the reference page,
    with the evidence outlined by the teal box; panels (b)--(d) show similarity
    maps for GARRET, ColQwen2.5, and RegionRAG, respectively. Each cross marks
    the peak similarity for that method. GARRET's peak falls inside the
    reference evidence, whereas both baselines peak outside it.}
    \label{fig:supp_similarity_pharma}
\end{figure*}

\begin{figure*}[p]
    \centering
    \includegraphics[width=\textwidth,height=0.80\textheight,
    keepaspectratio,trim=18bp 0 35bp 0,clip]
    {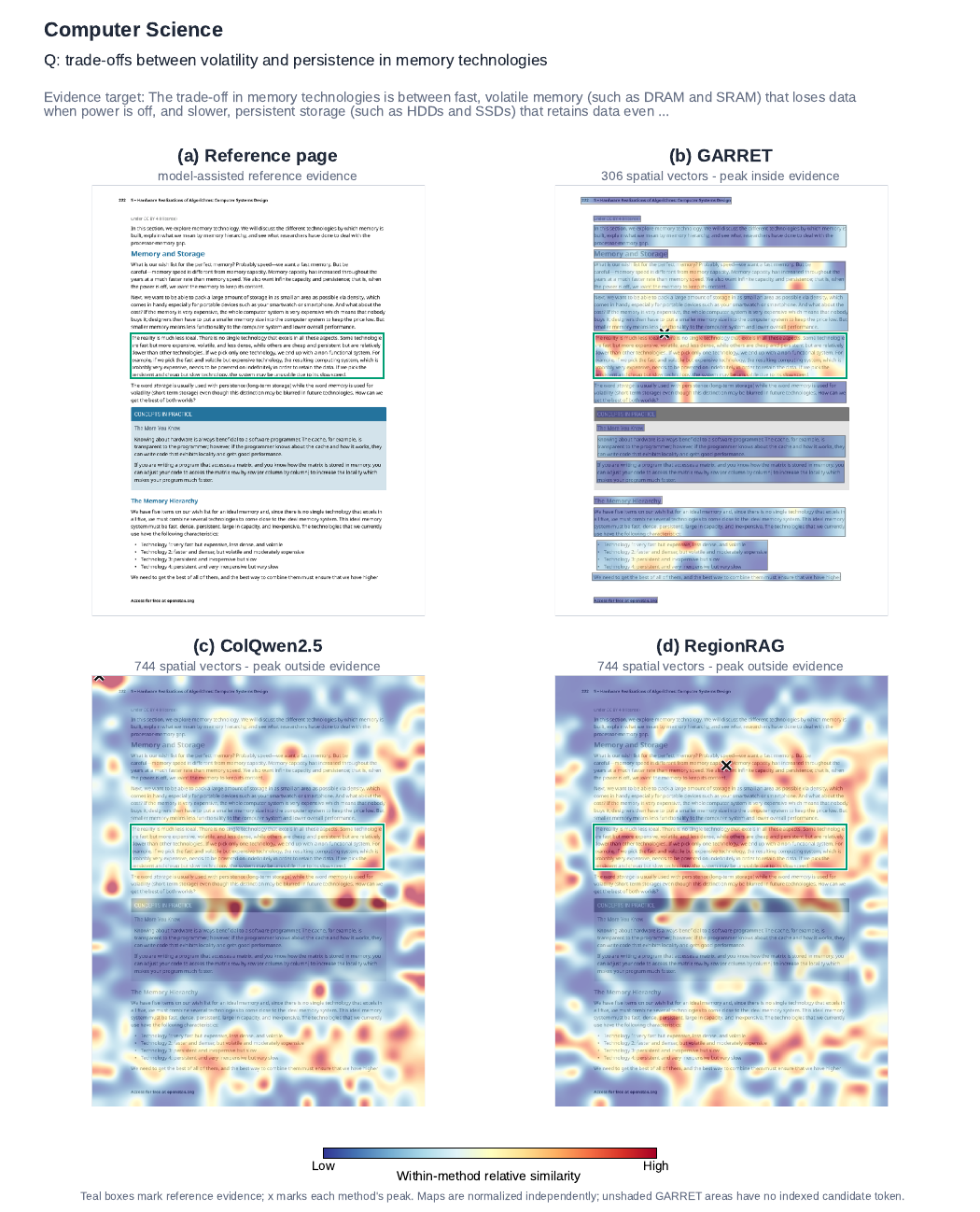}
    \caption{Query--page spatial similarity for an example from the
    computer-science subset of ViDoRe V3. Panel (a) shows the reference page,
    with the evidence outlined by the teal box; panels (b)--(d) show similarity
    maps for GARRET, ColQwen2.5, and RegionRAG, respectively. Each cross marks
    the peak similarity for that method. GARRET's peak falls inside the
    reference passage, whereas both baselines peak elsewhere.}
    \label{fig:supp_similarity_cs}
\end{figure*}

Figures~\ref{fig:supp_similarity_pharma}
and~\ref{fig:supp_similarity_cs} show query-to-spatial-token similarity over
complete pages. These are retrieval-similarity maps; decoder attention uses a
separate signal. Teal boxes mark the reference evidence, and a cross marks
each retriever's peak response. We normalize colors independently within each
method. Colors therefore show spatial concentration and peak location within
one method, and each method uses a separate color scale. Unshaded areas in the
GARRET panels contain no indexed candidate token. In both examples, GARRET's
peak lies inside the reference evidence, while the other retrievers peak
elsewhere.
\clearpage
\begin{table*}[p]
    \centering
    \begingroup
    \setlength{\fboxsep}{6pt}
    \textbf{Evidence-to-Answer Generation Prompt}\\[4pt]
    \fbox{%
    \begin{minipage}{0.94\textwidth}
        \footnotesize\ttfamily\raggedright
        You are answering a question using only the provided document
        evidence.\\[3pt]
        Rules:\\
        1. Use only the provided evidence.\\
        2. If the evidence is insufficient, set "insufficient" to true and
        use the answer "INSUFFICIENT\_EVIDENCE".\\
        3. Preserve numbers, units, names, and dates exactly.\\
        4. Answer in the same language as the question.\\
        5. Give a concise answer, usually one or two short sentences.\\
        6. Cite the evidence ids that support the answer, such as E1 or E3.\\
        7. Return exactly one valid JSON object. Do not include markdown
        fences, prose, or reasoning outside JSON.\\[3pt]
        Question:\\
        \{query\}\\[3pt]
        Evidence:\\
        \{evidence\_text\}\\[3pt]
        Return JSON:\\
        \{\\
        \hspace*{1em}"answer": "...",\\
        \hspace*{1em}"evidence\_ids": ["E1"],\\
        \hspace*{1em}"insufficient": false\\
        \}
    \end{minipage}}\\[9pt]
    \textbf{Answer-Correctness Judge Prompt}\\[4pt]
    \fbox{%
    \begin{minipage}{0.94\textwidth}
        \footnotesize\ttfamily\raggedright
        Act as a strict answer-quality judge.\\[3pt]
        Compare the Test Answer with the Reference Answer while considering
        the Question. Judge whether the test answer\\
        preserves the essential facts needed to answer the question.
        Evaluate meaning rather than wording.\\[3pt]
        Use exactly one label:\\
        - Correct: all essential information is present and accurate. Small
        omissions of nonessential detail and small,\\
        \hspace*{1em}relevant additions are acceptable when they do not
        conflict with the reference.\\
        - Partially Correct: some essential information is right, but
        important content is missing, or the answer adds\\
        \hspace*{1em}substantial unrequested material.\\
        - Incorrect: the answer is fundamentally wrong, contradicts the
        reference, or fails to provide the essential answer.\\[3pt]
        Question:\\
        \{query\}\\[3pt]
        Reference Answer:\\
        \{reference\_answer\}\\[3pt]
        Test Answer:\\
        \{model\_answer\}\\[3pt]
        Return exactly one JSON object and no markdown:\\
        \{\\
        \hspace*{1em}"explanation": "brief justification",\\
        \hspace*{1em}"judgment":
        "Correct|Partially Correct|Incorrect"\\
        \}
    \end{minipage}}
    \endgroup
    \caption{Exact prompts for the downstream RAG evaluation on ViDoRe V3.
    The first prompt generates an evidence-grounded answer, and the second
    assigns the strict correctness label used to compute accuracy. Braced
    fields are instantiated per query.}
    \label{tab:supp_rag_prompts}
\end{table*}
\clearpage
\begin{table*}[p]
    \centering
    \begingroup
    \setlength{\fboxsep}{6pt}
    \textbf{Region-Relevance Scoring Prompt}\\[4pt]
    \fbox{%
    \begin{minipage}{0.94\textwidth}
        \footnotesize\ttfamily\raggedright\linespread{0.89}\selectfont
        You are an expert document analysis AI. You will be provided with an
        image of a document that has a specific area highlighted with a RED
        bounding box. You will also be given a user query.\\[3pt]
        Your task is to determine how relevant the information located
        strictly inside the RED bounding box is to answering the user's
        query.\\[3pt]
        IMPORTANT RULES:\\
1. You must ONLY evaluate the contents enclosed by the red box.\\
2. You must not use information located outside the red box to justify the
        score, even if the correct answer appears elsewhere on the same
        page.\\
3. A region should receive the highest score only when the content inside the
        red box provides direct or indispensable evidence for answering the
        query.\\
4. Incidental geometric overlap, topical similarity, shared keywords, or
        relevant evidence located primarily outside the red box are not
        sufficient for the highest score.\\
5. Assign exactly one integer score from 0 to 5.\\[3pt]
        Please evaluate the match between the query and the red box contents
        using the following 0-5 scoring system:\\[3pt]
        Score 5 (Direct or Indispensable Evidence): The content inside the red
        box directly, accurately, and sufficiently answers the query, or
        provides evidence that is indispensable for deriving the answer. The
        essential evidence is primarily contained inside the red box and
        requires little or no information from outside the box.\\[3pt]
        Score 4 (Highly Relevant but Insufficient): The content inside the red
        box is highly relevant and provides a substantial part of the answer,
        but it is not sufficient on its own. An essential detail is missing,
        the answer depends materially on information outside the box, or the
        box contains only one component of a multi-part answer.\\[3pt]
        Score 3 (Relevant Supporting Evidence): The content inside the red box
        provides useful supporting information for answering the query, but
        it addresses only part of the query or provides indirect evidence. It
        does not constitute a direct answer or indispensable evidence by
        itself.\\[3pt]
        Score 2 (Topically Relevant): The content inside the red box discusses
        the same topic, entity, event, attribute, or concept as the query, but
        it does not provide meaningful evidence for answering the query.\\[3pt]
        Score 1 (Superficial Match): The content inside the red box contains
        isolated keywords, entities, numbers, or visual elements associated
        with the query, but the connection is incidental and the content does
        not substantively address the query.\\[3pt]
        Score 0 (No Match): The content inside the red box is unrelated to the
        query and provides no useful information for answering it.\\[3pt]
        
        Output Format Requirement:\\
        You must output your response STRICTLY as a valid JSON object. Do not
        include markdown formatting, code fences, conversational text, or
        explanations outside the JSON object.\\[3pt]
        The JSON object must contain exactly two keys:\\
- "reasoning": A string containing 1-2 sentences explaining why the content
        inside the red box does or does not provide evidence for answering the
        query.\\
- "score": An integer representing the score: 0, 1, 2, 3, 4, or 5.\\[3pt]
        Example output:\\
        \{\\
        \hspace*{1em}"reasoning": "The boxed content directly states the
        requested value and identifies the relevant entity, so it fully
        answers the query without relying on information outside the box.",\\
        \hspace*{1em}"score": 5\\
        \}\\[3pt]
        User Query:\\
        \{query\}
    \end{minipage}}\\[9pt]
    \endgroup
    \caption{Exact Qwen3-VL prompt used for VLM-assisted relevance
    pre-annotation in \textsc{ViDoRe-V3-Region}. Each input highlights one
    candidate in red, and the model must judge only the enclosed content and
    return a short rationale with an integer relevance score from 0 to 5.}
    \label{tab:supp_region_relevance_prompt}
\end{table*}

\clearpage

\clearpage

\clearpage

\clearpage

\clearpage
\end{document}